\documentclass[twoside]{IEEEtran}
\usepackage{amsmath,amsfonts}
\usepackage{algorithm}
\usepackage{array}
\usepackage[caption=false,font=normalsize,labelfont=sf,textfont=sf]{subfig}
\usepackage{textcomp}
\usepackage{stfloats}
\usepackage{url}
\usepackage{verbatim}
\usepackage{graphicx}
\usepackage{cite}
\usepackage{adjustbox}
\usepackage{algorithm}
\usepackage{algorithmicx}
\usepackage{bm}
\usepackage{xcolor}
\usepackage{multirow}
\usepackage{graphics}
\usepackage{fancyhdr}
\usepackage{authblk}

\IEEEoverridecommandlockouts
\usepackage{tikz}
\usepackage{textcomp}
\usepackage{hyperref}
\usepackage{lipsum}

\newcommand\copyrighttext{%
  \footnotesize \textcopyright 2022 IEEE. Personal use of this material is permitted. Permission from IEEE must be obtained for all
other uses, in any current or future media, including reprinting/republishing this material for advertising
or promotional purposes, creating new collective works, for resale or redistribution to servers or lists, or
reuse of any copyrighted component of this work in other works.}
\newcommand\copyrightnotice{%
\begin{tikzpicture}[remember picture,overlay]
\node[anchor=south,yshift=5pt] at (current page.south) {\fbox{\parbox{\dimexpr\textwidth-\fboxsep-\fboxrule\relax}{\copyrighttext}}};
\end{tikzpicture}%
}

\begin{document}

\title{Knowledge-augmented Deep Learning and Its Applications: A Survey}

\author{Zijun Cui, Tian Gao, Kartik Talamadupula, and Qiang Ji
\thanks{Z. Cui and Q. Ji are with the Department of Electrical, Computer, and Systems Engineering, Rensselaer Polytechnic Institute, Troy,
NY 12180, USA (e-mail: cuiz3@rpi.edu; jiq@rpi.edu).}
\thanks{T. Gao and K.Talamadupula are with IBM Research AI, USA.}
}


\markboth{IEEE Transactions on Neural Networks and Learning Systems}{CUI \MakeLowercase{\textit{et al.}}: Knowledge-augmented Deep Learning and Its Applications}



\maketitle

\copyrightnotice

\begin{abstract}
Deep learning models, though having achieved great success in many different fields over the past years, 
are usually data hungry, fail to perform well on unseen samples, and lack of interpretability. Various prior knowledge often exists in the target domain and their use can alleviate the
deficiencies with deep learning. 
To better mimic the behavior of human brains, different advanced methods have been proposed to identify domain knowledge and integrate it into deep models for data-efficient, generalizable, and interpretable deep learning, which we refer to as \textit{knowledge-augmented deep learning} (KADL). 
In this survey, we define the concept of KADL, and introduce its three major tasks, i.e., knowledge identification, knowledge representation, and knowledge integration. Different from existing surveys that are focused on a specific type of knowledge, we provide a broad and complete taxonomy of domain knowledge and its representations. Based on our taxonomy, we provide a systematic review of existing techniques, different from existing works that survey integration approaches agnostic to taxonomy of knowledge. This survey subsumes existing works and  offers a bird’s-eye view of research in the general area of knowledge-augmented deep learning. The thorough and critical reviews of numerous papers help not only understand current progresses but also identify future directions for the research on knowledge-augmented deep learning.
\end{abstract}


\section{Introduction}
\IEEEPARstart{D}{espite} the impressive performance that existing deep models have achieved in various fields, they suffer from several serious deficiencies, including high data dependency and poor generalization~\cite{marcus2018deep}. These deficiencies originate primarily from the models' data-driven nature and their inability to effectively exploit the domain knowledge. To address these limitations, a knowledge-augmented deep learning paradigm begins to attract researchers' attention, whereby domain knowledge and observable data work together synergistically to produce data-efficient, generalizable, and interpretable deep learning algorithms.

Real-world domain knowledge is rich. In the context of deep learning, the domain knowledge mainly originates from two sources: target knowledge and measurement knowledge. Target knowledge governs the behaviors and properties of the target variables we intend to predict, while measurement knowledge controls the underlying mechanism that produces the observed data of the target variables. Based on its representation, we propose to divide the domain knowledge that has been explored in deep learning into two categories: \textit{scientific knowledge} and \textit{experiential knowledge}.
Scientific knowledge represents the well-established laws or theories in a domain that govern the properties or behaviors of target variables. 
In contrast, experiential knowledge refers to well-known facts or rules extracted from longtime observations and can also be inferred through humans' reasoning. 
Knowledge can be represented and organized in various formats. 
Scientific knowledge is usually well represented rigorously with mathematical equations.  Experiential knowledge, on the other hand, is usually represented less formally, such as through logic rules, knowledge graphs or probabilistic dependencies.  Knowledge with different representations is integrated with data in a deep learning framework through different integration approaches. 
\IEEEpubidadjcol

Recognizing the deficiencies with current deep learning, there is a growing interest recently in  capturing and encoding prior knowledge into the deep learning. 
Two mainstream techniques are neural-symbolic integration and physics informed deep learning.
Neural-symbolic integration models focus on encoding experiential knowledge into
the traditional symbolic AI models and integrating the symbolic models with deep learning models.
The physics informed deep learning focuses on encoding various
theoretical physical knowledge into different stages of deep learning. Current survey papers in this area are limited in scope as they focus on reviewing either neural-symbolic models or physics-informed machine learning methods, while ignoring many other
related  works. Specifically, the existing surveys on neural-symbolic models  mainly consist of the discussions on logic rules or knowledge graphs, and their integration into deep models~\cite{besold2017neural,yu2021survey}.  Existing surveys on physics-informed machine learning, however, are limited to a specific scientific discipline, and the integration methodologies are usually task-specific, e.g., physics~\cite{willard2020integrating, han2020integrating}, cyber-physical systems~\cite{rai2020driven}, geometry~\cite{rath2020boosting}, and chemistry~\cite{nowack2018using}. These surveys are hence focused on the methodologies for solving scientific problems under a lab environment, lack of discussions on real-world applications.
To address this limitation, 
we present a comprehensive yet systematic review of existing works on knowledge-augmented deep learning. The contributions of our survey are three folds:
\begin{itemize}
    \item This survey creates a novel taxonomy on domain knowledge, including both scientific knowledge and experiential knowledge. Our work subsumes existing works that focus on a subset of domain knowledge on specific disciplines~\cite{willard2020integrating,han2020integrating,rai2020driven,rath2020boosting,nowack2018using}.
    \item This survey covers a wide array of methodologies for knowledge representation and integration with a systematic categorization. 
    It differs from existing surveys on general integration techniques, that are agnostic to the taxonomy of domain knowledge~\cite{deng2020integrating, von2019informed,kim2021knowledge,von2020combining,sagel2020knowledge}. 
    
    \item This survey covers the methodologies that are not only for solving scientific problems under a lab environment, but, more importantly, for real-world application tasks. Not being limited to a specific application task, this survey involves the tasks ranging from computer vision to natural language processing.
    Our survey is hence of interest not only to deep learning researchers but also to 
    deep learning practitioners in different fields. 
\end{itemize}

We organize this survey as follows. We first introduce the concept of KADL in Section~\ref{sec:intro_knowledge-augmented_deep_learning}, whereby we define the three basic tasks (i.e., knowledge identification, knowledge representation, and knowledge integration). We then review KADL methodologies, where we categorize different techniques based on their domain knowledge of focus: 1) deep learning with scientific knowledge in Section~\ref{sec:scientific_knowledge} and 2) deep learning with experiential knowledge in Section~\ref{sec:experiential_knowledge}. In each category, we identify the domain knowledge, its representation formats, and the existing methods proposed for the integration of knowledge with data. An overview of existing methodologies in knowledge-augmented deep learning is included in Table~\ref{tab:overall-summary}.

\section{Knowledge-augmented Deep Learning}
\label{sec:intro_knowledge-augmented_deep_learning}

Major tasks of knowledge-augmented deep learning include knowledge identification, knowledge representation, and knowledge integration into deep models.
In the following sections, we introduce each of the major tasks in detail.
\begin{table*}[ht!]
    \centering
    \caption{Overview of Knowledge-augmented Deep Learning}
    \label{tab:overall-summary}
    \scalebox{0.9}{
    \begin{tabular}{|c|c|c|c|c|c|c|c|}
    \hline
        \multirow{2}{*}{\textbf{Knowledge Identification}} &\multirow{2}{*}{\textbf{Knowledge Representation}} &\multicolumn{4}{|c|}{\textbf{Knowledge Integration}} \\ 
        && Data-level& Architecture-level& Training-level& Decision-level\\ \hline
        \multirow{7}{*}{Scientific Knowledge} &  \multirow{6}{*}{Mathematical Equations} &\cite{chen2013data}  &\cite{liu2021discovering}\cite{jaques2021newtonianvae}\cite{daw2020physics}\cite{muralidhar2020phynet} &\cite{beucler2019achieving} \cite{beucler2021enforcing}\cite{zhang2018deeppotential}\cite{seo2021controlling} &\\ 
        &&  &\cite{sun2020theory}\cite{toth2019hamiltonian}\cite{choudhary2020physics} &\cite{djeumou2021neural}\cite{raissi2019physics}\cite{karniadakis2021physics}\cite{li2019estimating} &\\ 
        &&  &\cite{greydanus2019hamiltonian} \cite{zhong2019symplectic} \cite{cranmer2020lagrangian}\cite{allen2020lagnetvip}&\cite{stewart2017label}\cite{xie2020deep} \cite{wu2017learning} \cite{qiao2021differentiable} &\\ 
        &&  &\cite{botev2021priors} \cite{wang2020incorporating}\cite{finzi2021residual}\cite{ling2016reynolds} &\cite{mezghanni2021physically}\cite{shah2019encoding}\cite{wu2020enforcing}\cite{geneva2020modeling} &\\ 
        &&& \cite{anderson2019cormorant} \cite{lengyel2021zero} \cite{chang2016compositional}\cite{battaglia2016interaction}&\cite{takeishi2021physics,yin2021augmenting, linial2021generative}\cite{yang2019adversarial}&\\ 
        &&&&\cite{zhu2019physics} \cite{karumuri2020simulator}&\\ 
        \cline{2-6}
         &  \multirow{4}{*}{Simulation Engines} &\cite{jia2019physics}\cite{jia2021physics}\cite{bousmalis2018using}\cite{shah2018airsim} &&&  \\ 
         && \cite{coumans2016pybullet}\cite{ehsani2020use} \cite{mottaghi2016newtonian}\cite{tobin2017domain}&&&  \\ 
         && &&&  \\ 
         && &&&  \\ \hline
        \multirow{6}{*}{Experiential Knowledge} &  \multirow{2}{*}{Probabilistic Relationships} &\cite{chen2013data}\cite{li2013data}\cite{teshima2021incorporating} &\cite{Corneanu2019}\cite{chen2019deep}\cite{suhail2021energy}\cite{song2021hybrid}&\cite{srinivas2021matching}\cite{rieger2020interpretations}\cite{zhang2018classifier}\cite{cui2020knowledge}&\cite{li2013data}\cite{chen2013data}\\
        & & &\cite{yang2021causalvae}&&\\
        & & &&&\\
        & & &&&\\\cline{2-6}
         &  \multirow{2}{*}{Logic Rules} & \cite{seo2021controlling}&\cite{qu2019probabilistic}&\cite{xu2018semantic}\cite{xie2019embedding}&\\
         & & &&&\\
         & & &&&\\\cline{2-6}
         &  \multirow{2}{*}{Knowledge Graphs} & &\cite{liang2018symbolic}\cite{liu2021auto}\cite{luo2020context}&\cite{fang2017object}\cite{gu2019scene}&\cite{marino2021krisp}\cite{cui2021type}\\
         & & &&&\\
         & & &&&\\\hline
    \end{tabular}}
\end{table*}

\subsection{Knowledge Identification} 
Knowledge identification involves identifying the domain knowledge for a specific task. For some tasks, domain knowledge is readily available and hence easy to identify, while in other tasks, the knowledge is less straightforward and requires domain experts' efforts to identify. 
In this survey, we divide the domain knowledge into \textit{scientific knowledge} and \textit{experiential knowledge}. 
Scientific knowledge is prescriptive and mainly refers to well-formulated mathematical theories or physics laws. These laws are validated extensively through scientific experiments and are true in a universal setting (e.g., Newton's laws). These laws explicitly define the deterministic relationships among different variables that are involved in the system. 
Recent work in physics-informed machine 
learning seeks to exploit various scientific knowledge from different disciplines for enhancing deep learning.
Experiential knowledge is descriptive and mainly refers to well-known facts from daily life, indicating semantic properties of an entity or semantic relationships among multiple entities. Experiential knowledge usually is extracted from longtime observations but can also be derived from well-established studies or theories. The latter type of experiential knowledge is science-grounded and is focused on a semantic and abstract level of description. Experiential knowledge usually contains lots of fragmented information, and could be uncertain, imprecise, or ambiguous.  Recent work in neural-symbolic models focuses on embedding experiential knowledge into deep learning.


\subsection{Knowledge Representation}
Knowledge representation involves representing the identified domain knowledge in a well-organized and structured format. 
The appropriate representation depends on the type of domain knowledge. Scientific knowledge is usually expressed using equations.
Besides, a simulation engine is also considered as an alternative representation of the scientific knowledge. 
Experiential knowledge is less formal compared to scientific knowledge. Experiential knowledge can be represented through probabilistic relationships, logic rules or knowledge graphs. 

\subsection{Knowledge Integration}
\begin{figure}[ht!]
    \centering
    \includegraphics[width=2.5in, height=3in]{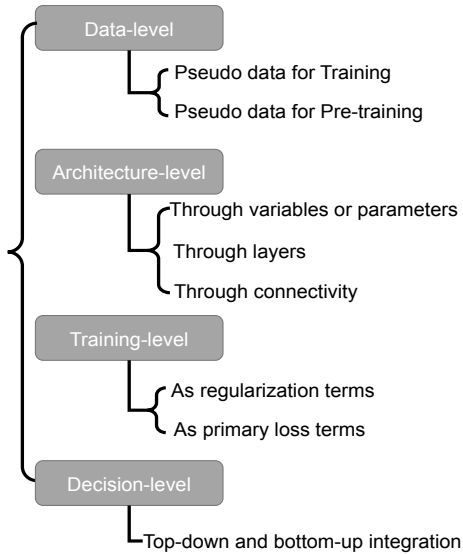}
    \caption{Four levels of integration: data-level, architecture-level, training-level, and decision-level.}
    \label{fig:integrationlevels}
\end{figure}

Knowledge integration entails integrating domain knowledge into deep models. Through the integration, a deep model can leverage both existing datasets and domain knowledge for certain tasks. Different integration methods can be employed depending on the types of knowledge, and can be divided into four levels: data-level, architecture-level, training-level, and decision-level, as shown in Figure~\ref{fig:integrationlevels}. 
Data-level integration focuses on integrating the knowledge by augmenting the original training data with pseudo data, which is generated based on the knowledge. Architecture-level integration embeds knowledge by modifying neural network architecture. Training-level integration regularizes the training of a deep model via regularization terms or primary loss terms that are derived from the knowledge. Finally, decision-level integration combines a top-down prediction from a prior model with a bottom-up prediction from a deep model, whereby knowledge encoded in the prior model help refine the predictions of a deep learning pipeline.

Each type of integration could be beneficial from different aspects. Data-level integration can help alleviate data paucity issue that faced by many deep models~\cite{jia2019physics, jia2021physics, bousmalis2018using, shah2018airsim}. Additionally, data generation via automatic simulation is usually less expensive compared to human annotation. Architecture-level integration brings the bonus of making deep models interpretable and explainable, which are two crucial factors for trustworthy artificial intelligence~\cite{kaur2022trustworthy}. Training-level integration can be regarded as the most common approach due to its straightforwardness. It can be flexibly applied to different deep model frameworks. 
Flexibility of training-level integration also promotes the quantification of knowledge uncertainties~\cite{yang2019adversarial, zhu2019physics, geneva2020modeling, karumuri2020simulator}. Decision-level integration, 
different from previous three approaches, employs knowledge independent of the training of deep models, and is relatively less explored by existing works.

Selection of integration approach could be both task-dependent and knowledge-dependent. Architecture-level integration would be preferred if the knowledge is expected to be integrated in a hard way. The rest of three integration approaches (i.e., data-level, training-level, or decision-level) introduce knowledge into a deep learning pipeline in a soft way. Secondly, if the knowledge only involves target variables (i.e., predictions of a neural network), training-level integration would be preferred. To perform the other three types of integration, the knowledge is expected to involve measurements (e.g., intermediate variables or observations) and target variables. 
Lastly, if the identified knowledge consists of highly nonlinear and complex relationships, leveraging well-established engines or simulators for data-level integration would be the primary choice. 

\section{Deep Learning with Scientific Knowledge}
\label{sec:scientific_knowledge}
Deep learning models are gaining importance in advanced science and engineering areas that have traditionally been dominated by mechanistic (e.g., first principle) models. Such models produce particularly promising performances for scientific problems whose undergoing mechanisms are not well understood by experts or those problems for which exact solutions are computationally infeasible.  Existing deep learning, however, requires a significant amount of annotated data, and it generalizes poorly to novel data or settings. 

There is a growing consensus in the research community regarding combining conventional methodologies in science and engineering with existing data-driven deep models. 
Deep learning with scientific knowledge explores the continuum between classical mechanistic models and modern deep ones.
There have been growing efforts in the machine learning community to incorporate scientific knowledge into deep learning (also referred to as physics-informed machine learning) to generate physically consistent and interpretable predictions and to reduce data dependency.  

In the following, we first identify the types of scientific knowledge and their representations. 
We then introduce different methodologies on integrating scientific knowledge  with deep models.

\subsection{Scientific Knowledge Identification} 
Scientific knowledge refers to well-formulated mathematical or physics equations that have been validated through extensive scientific experiments and are true in a universal setting (e.g., Newton's laws). These laws explicitly define a deterministic and precise relationship among different objects that are involved in a system.

Current physics-informed deep learning seeks to explore the usage of classical mechanics models. For a dynamic system, the most widely considered scientific knowledge is Newtonian mechanics, which includes kinematics and dynamics. The former refers to observable motion (such as motion trajectories), which is typically represented with polynomial equations involving measurable properties (e.g., velocity, acceleration, or positions). 
Kinematics studies motion without regard for the cause. In contrast, dynamics studies the causes of a motion, whereby partial differential equations (PDEs) are used to capture the relationships between forces and measurable properties. Existing works explore the usage of dynamics in various physics systems (e.g., gas and fluid dynamics~\cite{wang2020physics} and molecular dynamics of protein~\cite{sultan2018transferable, unke2021se}).  With the understanding of dynamics, kinematics can be better predicted. Newtonian mechanics has hence been leveraged in real-world applications, such as human body behavior analysis~\cite{li2019estimating, yuan2021simpoe,shimada2021neural}. 
Newtonian mechanics, unfortunately, can result in equations of motion that are intractable to solve, even for a seemingly simple system (e.g., double pendulum system). \textit{Lagrangian mechanics} or \textit{Hamiltonian mechanics} can instead be considered. 
As re-formulations to Newtonian mechanics, both Lagrangian mechanics and Hamiltonian mechanics leverage generalized coordinates, making them flexible with respect to which coordinates to use to understand a system. In Lagrangian mechanics, $L$ is defined as the difference between the kinetic energy, $T$, and potential energy, $U$, of a system ( i.e., $L = T - U$). The Hamiltonian $H$ resembles the Lagrangian $L$, and is defined as the summation of the kinetic energy, $T$, and potential energy, $U$, of a system (i.e., $H = T + U$). In Lagrangian mechanics, the time derivative of position $\dot{x}$ is considered as generalized momentum, whereas in Hamiltonian mechanics, momentum $p$ is considered. For simple particle systems, such differences are trivial, while in more complicated systems (e.g., magnetic fields), momentum can no longer be computed as simple a product of mass and velocity. 
The dynamic equations for both a Lagrangian system %
and a Hamiltonian system 
conserve energy over time with conservative forces.

Symmetries have also been widely explored in physics. Philip Anderson famously argued, ``It is only slightly overstating the case to say that physics is the study of symmetry''~\cite{anderson1972more}. 
Discovering symmetries has been proven to be important in both deepening the understanding of physics and enhancing machine learning algorithms. 
Equivariant or invariant functions preserve symmetries and have often been exploited for incorporating these symmetries into deep learning algorithms. 

Optics, another type of physics knowledge, has also been considered. Optics studies the behavior and properties of light. 
Fermat's principle~\cite{born2013principles} is the basic law in optics. 
In addition, the illumination models~\cite{oren1995generalization} and the rendering equation~\cite{kajiya1986rendering} capture the 3D object 
appearances with their image appearances. 
Existing works explore the usage of various illumination models for different computer vision tasks~\cite{cao2018sparse,wang2003illumination} and computer graphics tasks~\cite{xin2019theory, kim2020single, zhou2021learning}.

Besides physics knowledge, mathematical theories such as 
theorems on existing algorithms (e.g., sorting or ranking), as well as continuous relaxation~\cite{petersen2021learning}, have also been considered.  Projective geometry theories~\cite{hartley2003multiple} are
widely applied to various computer vision tasks.

\subsection{Representation of Scientific Knowledge}
\begin{figure*}[ht!]
    \centering
    \includegraphics[width=5.8in, height=3in]{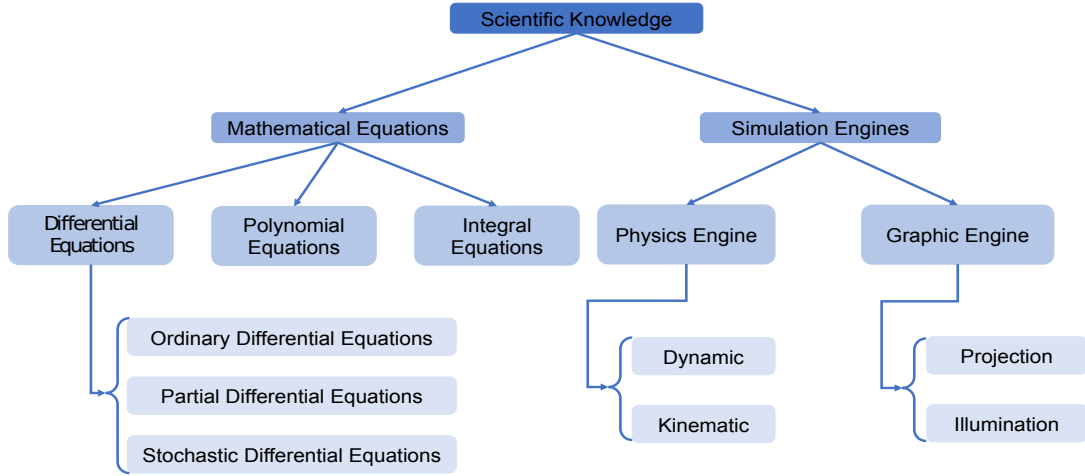}
    \caption{Taxonomy of scientific knowledge and its representations.} 
    \label{fig:taxonomy-scientific}
\end{figure*}
Scientific knowledge is usually represented as equations, such as differential equations. Another important representation tool of scientific knowledge is simulation. Simulation mimics a real-world physics system and is treated as a surrogate representation of the physics principles that govern the real-world physics system. We visualize the taxonomy of scientific knowledge and its representations in Figure~\ref{fig:taxonomy-scientific}.

\subsubsection{Mathematical equation} Equations can include polynomial equations, differential equations, and integral equations, among which differential equations are widely explored by existing works. 
Dynamic laws are usually represented by PDEs, which indicate a deterministic relationship among different variables. 
In general, the PDEs are of the form
\begin{equation}
    \begin{split}
        &\mathcal{N}_{\bm{x}}(u (\bm{x}); \bm{\lambda}(\bm{x})) = f(\bm{x}), \quad \bm{x} \in D\\
        &\mathcal{B}_{\bm{x}}(u(\bm{x}); \bm{\lambda}(\bm{x})) = b(\bm{x}), \quad \bm{x} \in \Gamma
    \end{split}
    \label{eq:pde}
\end{equation}
where $\bm{x}=\{x_1, x_2, ..., x_N\}$ are $N$ variables involved in the system. $\mathcal{N}_{\bm{x}}$ is a general differential operator, and $\mathcal{B}_{\bm{x}}$ is the boundary condition operator. $D$ represents a physical domain, and $\Gamma=\partial D$ represents the boundary of the domain. $\bm{\lambda}(\bm{x})$ represents the physics parameters involved in the PDEs. 
For simple systems, physical parameters are constants (i.e., $\bm{\lambda}(\bm{x})=\bm{\lambda}$). $f(\bm{x})$ is a forcing term, and $b(\bm{x})$ specifies the boundary condition, e.g., Dirichlet boundary conditions for the Darcy flow problem~\cite{zhu2019physics}. $u(\bm{x})$ is the solution of the differential equation given the specified boundary conditions.
The equations become ordinary differential equations (ODEs) when only one variable is present.
When terms (e.g., physics parameters $\bm{\lambda}$) exist in differential equations that undergo stochastic processes,  the equations become stochastic differential equations (SDEs). The general form of SDEs resembles standard differential equations, except for the random event $\omega$:
\begin{equation}
    \begin{split}
        &\mathcal{N}_{\bm{x}}(u(\bm{x};\omega); \bm{\lambda}(\bm{x};\omega)) = f(\bm{x};\omega), \quad \bm{x} \in \mathcal{D}, \quad \omega \in \Omega\\
        &\mathcal{B}_{\bm{x}}(u(\bm{x};\omega); \bm{\lambda}(\bm{x};\omega)) = b(\bm{x};\omega), \quad \bm{x} \in \Gamma
    \end{split}
    \label{eq:sde}
\end{equation}
The physics parameters $\bm{\lambda}(\bm{x};\omega)$ and forcing terms $f(\bm{x};\omega)$ are modeled as random processes, and thus the solution $u(\bm{x};\omega)$ follows stochastic processes specified by $f$ and $\bm{\lambda}$. 

Differential equations describe a system's evolution over time when we explicitly set one of the variables to correspond to time $t$, which commonly occurs in different dynamic systems. For example, the Euler-Lagrange equation defines the dynamics of a Lagrangian system,
\begin{equation}
    \frac{\partial L}{\partial x} - \frac{d}{d t}\frac{\partial L}{\partial \dot{x}} = 0
    \label{eq:lagrangian}
\end{equation}
which connects the derivatives of the Lagrangian $L$ with respect to location in generalized coordinates $x$, time $t$, and generalized momentum $\dot{x}$. Differential equations have been widely explored by existing works, such as Newton's second law~\cite{jaques2021newtonianvae, stewart2017label},  Burgers' equation occurring in gas and fluid dynamics~\cite{wang2020physics}, Hamilton's equations in Hamiltonian dynamics~\cite{choudhary2020physics},  Euler-Lagrange equations for Lagrangian dynamics~\cite{lutter2019deep}, and the Lorenz equations, which describe a nonlinear chaotic system used for atmospheric convection~\cite{satorras2019combining}. 

Laws of light are also expressed as equations. Fermat's principle considers the integration over a light path.  In~\cite{xin2019theory}, the transient $I$ in a transient imaging system is characterized through an integration as 
\begin{equation}
    I(\tau; \bm{v}) = \int_{\bm{x}\in \chi} f(\bm{x};\bm{v}) \delta(\tau - \tau({\bm{x};\bm{v}})) dA(p,q)
\end{equation}
where $\tau$ is the traveled pathlength and $\bm{v}$ is the visible point. $A(p,q)$ measures the unit area of a surface, and the function $f$ absorbs
reflectance and shading. The rendering equation represented as an integral equation has also been considered~\cite{Zhang_2022_CVPR}.
Equality algebraic equation has also been considered.  According to the law of reflection, 
an image $I$ with reflections is a sum of the glass reflected back scene $\hat{R}$ and the glass transmitted front scene $\hat{T}$, i.e.,
\begin{equation}
    I(x,y) = \hat{T}(x,y) + \hat{R}(x,y)
\end{equation}
Such equality algebraic equation is demonstrated to be helpful in reflection removal tasks~\cite{kim2020single}.
Another example is Malus' law, expressed as an algebraic equation, which defines the effect of polarization~\cite{zhou2021learning}.  


\subsubsection{Simulation engines} Besides explicitly representing physics laws with equations, simulation through engines is another method of representation. Simulation mimics a real physical system governed by physics laws and is hence considered a surrogate representation of the knowledge. 
A physics engine mainly encodes the governing dynamic laws of a physics system, such as rigid-body, soft-body, and fluid. It computes the accelerations, velocity, and displacement of an object from forces, by solving the equation of motion. Specifically, a physics engine simulates observable kinematics given certain causes following governing dynamic rules. To simulate a sequence of motion of rigid-body objects given specific forces, a physics engine is considered~\cite{ehsani2020use, zeng2020tossingbot, mottaghi2016newtonian, mottaghi2016happens, wu2017learning,wu2015galileo}. Most of these simulators are non-differentiable, making them prohibited to be employed in an end-to-end deep learning framework. There also exist simulation engines that encode kinematic laws for robotic manipulation. These inverse engines, however, estimate control actions using kinematic equations such that a desired position can be reached, agnostic to the underlying dynamic laws~\cite{xie2020deep}. 
Graphic engines have also been explored. A graphic engine encodes the principled projection and illumination models, and renders realistic 2D observations by following the governing principles. For example, an engine, governed by laws of reflection, was proposed for the generation of faithful image rendering~\cite{kim2020single}.

\subsection{Integration into Deep Models}
To integrate domain-specific scientific knowledge into deep models, existing methodologies can be classified into three categories: data-level, architecture-level, and training-level integration  as shown in Figure~\ref{fig:integrationlevels}. Decision-level integration is rarely considered for scientific knowledge. Below we review methods to integrate scientific knowledge into deep models using these methods.

\subsubsection{Data-level Integration} 
One way to harness domain knowledge is to train deep models with data synthetized from conventional mechanistic models. Physics-based mechanistic models capturing the domain knowledge serve as simulators, and are used for generating synthetic data. The simulated data can either be combined with real data to jointly train the model or be used independently to pre-train the model through self-supervised learning.

\begin{figure}[ht!]
    \centering
    \includegraphics[width = 3in]{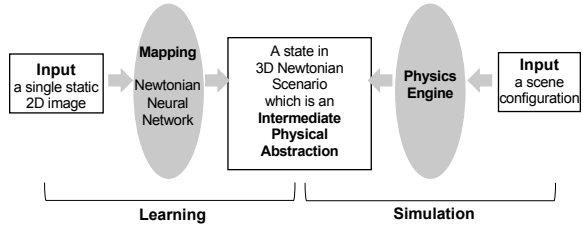}
    \caption{Understanding physics from 2D images through simulation~\cite{mottaghi2016newtonian}. The game engine takes scene configurations as input and generates in total 66 Newtonian scenarios as an abstraction of the real physical world. The Newtonian neural network learns to understand the physics by classifying a static image into one of the 66 possible Newtonian scenarios.}
    \label{ch2-relatedworks-newtonianimage}
\end{figure}

Simulated data has been widely employed for entire training, where no additional real training data is required. In computer vision, Mottaghi et al.~\cite{mottaghi2016newtonian} proposed understanding the force acting on a query object by predicting its long-term motion in 3D space as the response to the force, given a static 2D image. An overview of its procedure appears in Figure \ref{ch2-relatedworks-newtonianimage}.
The Blender\footnote{http://www.blender.org/} game engine, consisting of a physics and a graphic engine, is employed for the simulation. 
Specifically, the physics engine takes a scene configuration as input (e.g., a ball sliding on a slide) and simulates the scene forward in time according to motion equations derived from Newton's second law to produce different Newtonian scenarios. Physics parameters (e.g., force magnitude) are randomly sampled. 
For each Newtonian scenarios, the graphic engine renders 2D videos from the synthetic 3D world under different viewpoints in perspective projection. 
In total, 66 synthetic videos corresponding to 12 possible physical generated scenarios are used for training. Similarly, to better understand the human interaction with objects from 2D videos, Ehsani et al. \cite{ehsani2020use}  proposed a physical understanding of actions by inferring contact points together with forces from videos. A forward physics simulation is applied to supervise the force estimation, given observed videos, without requiring GT labels for forces. In particular, the 3D contact points of a moving object are estimated over a period given estimated forces, through physics simulation which is governed by Newton's second law. A projection operator is then applied transforming the estimated 3D keypoints into a 2D space. By minimizing the difference between the estimated and observed 2D contact points over time, physically consistent forces are obtained. To perform differentiable physics simulations, a finite difference method is applied for gradient computation using the PyBullet\footnote{https://pybullet.org/wordpress/} simulator, which is focused on rigid-body simulation following Newton's second law. In a similar vein, Tobin et al.~\cite{tobin2017domain} showed the effectiveness of synthetic samples generated by robotic simulation for training a deep model for object localization tasks, which is instrumental in robotic manipulation. Simulation is performed by employing the MuJoCo physics engine,\footnote{https://mujoco.org/}, whereby simulated 2D images are generated based on its built-in graphic engine~\cite{todorov2012mujoco}.
In particular, the MuJoCo physics engine is built based on Newtonian mechanics. The built-in graphic engine renders a 2D image given a selected camera in the 3D virtual environment through a perspective projection. To ensure sufficient simulation variabilities, a domain randomization strategy is proposed, where simulation parameters, such as the positions and orientation of objects, are all randomly specified during the simulation. 

Deep model training often starts with a pre-training stage, followed by a fine-tuning. 
Existing studies also show that pre-training affects the final performance of deep models, mainly because poor pre-training can lead models to anchor in a local optimum. Pre-training through simulated data has shown to help improve parameters' initialization. Jia et al.~\cite{jia2019physics, jia2021physics} introduced a physics-guided recurrent neural network (PGRNN) to model lake temperature dynamics. 
PGRNN is pre-trained on synthetic data generated from a physics-based mechanistic simulator and is then fine-tuned with some observation data. 
The simulator models the lake temperature dynamics as a function of physics parameters (e.g., water clarity and wind sheltering) through PDEs. It was shown that, even with synthetic data generated with an imperfect set of physics parameters, PGRNN still achieves competitive performance.  
Such an idea has also been explored in engineering disciplines. In robotics, Bousmalis et al.~\cite{bousmalis2018using} showed that the observation data required for accurate object grasping is significantly reduced (by a factor of 50) through physics-guided initialization. In autonomous driving, Shah et al.~\cite{shah2018airsim} pre-trained a driving algorithm with synthetic samples generated by a proposed simulator built on a game engine with physics laws embedded. In particular, the simulator includes a vehicle model and a physics engine. To describe a virtual 3D environment, physics parameters, such as gravity, air density, air pressure, and magnetic field, are specified manually. With these specified parameters,  the physics engine predicts the kinematic states, given forces and torques estimated from a vehicle model by following motion equations derived from Newton's second law.
This work showed that the data needs of driving algorithms can be drastically lessened through pre-training with simulated samples. In addition to augmenting data through simulation with physics engines, synthetic data can be generated from mathematical equations~\cite{chen2013data}.

\subsubsection{Architecture-level Integration} 
Domain knowledge can also be integrated through a customized design of neural network architectures. 
Architecture-level integration can be accomplished by 1) introducing specific physically meaningful variables or neural network parameters, 2) introducing layers derived from domain knowledge,  and 3) introducing physics-inspired connectivity among neurons. We introduce each type of approaches at below. 

\begin{figure}[ht!]
    \centering
    \includegraphics[width=3in, height=1.5in]{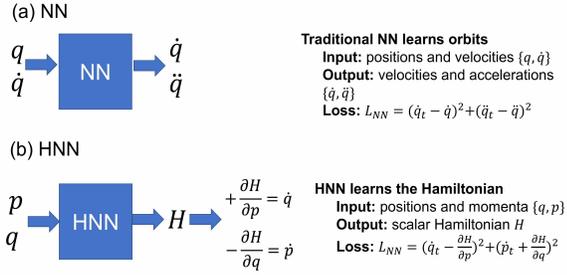}
    \caption{Comparison between a Hamiltonian neural network (HNN) and conventional neural network (NN). For dynamic modeling, instead of directly predicting trajectories, the HNN predicts the Hamiltonian $H$ of a system, through which the trajectory can be obtained under the energy conservation constraint.} 
    \label{fig:ch2-relatedworks-hamiltonian}
\end{figure}

\paragraph{Integrating through variables or parameters} 
One way to embed physical principles into the architecture of neural networks is to introduce physically meaningful variables in neural networks. The variable can be the output node of a neural network. Hamiltonian functionality that enforces energy conservation has attracted much attention~\cite{choudhary2020physics, toth2019hamiltonian, greydanus2019hamiltonian, zhong2019symplectic}. The Hamiltonian operator in physics is the
primary tool for modeling dynamic systems with conserved quantities.  
In Hamiltonian mechanics, a classical physical system is described by a generalized coordinate, $q$, and a conjugate momenta, $p$.
Hamiltonian $H$ then calculates the total energy of the system. The Hamiltonian equations defining the dynamics of a system are as follows:
\begin{equation}
    \dot{q} = \frac{\partial H}{\partial p}; \quad \dot{p} =- \frac{\partial H}{\partial q}
    \label{eq:hamiltonian}
\end{equation}
Inspired by the Hamiltonian mechanics, a Hamiltonian neural network (HNN) has been proposed, where the output represents the Hamiltonian dynamics, through which energy conservation is explicitly enforced~\cite{toth2019hamiltonian}. 
The differences between traditional NNs and HNNs can be understood easily, as shown in Figure~\ref{fig:ch2-relatedworks-hamiltonian}, which shows that traditional NNs learn to predict particle trajectories, while HNNs learn a particle's Hamiltonian, upon which trajectories can be predicted.
Choudhary et al.~\cite{choudhary2020physics} later showed that an HNN can better handle highly nonlinear dynamics in a chaos system by enforcing the conservation of the total energy. 
To demonstrate the practical importance of Hamiltonian formalism, Hamiltonian functionality was incorporated into a generative network, where Hamiltonian dynamics is learned from 2D observations without domain coordinate assumptions, such as images. The proposed Hamiltonian generative network has been applied to density estimation, leading to a neural Hamiltonian flow~\cite{toth2019hamiltonian}. By leveraging Hamiltonian formalism, the total probability can be easily conserved, while the density modeling remains expressive. Sharing a similar idea with HNNs, a generalized energy conservation employing Lagrangian mechanics was explored in the Lagrangian neural network (LNN)~\cite{cranmer2020lagrangian, allen2020lagnetvip}, where the output of the LNN is the Lagrangian dynamics.  
Nevertheless, evaluations of all these existing proposed models remain conceptual, without real applications in practice~\cite{botev2021priors}.

Variables can also be intermediate variables in a neural network. 
Jaques et al.~\cite{jaques2021newtonianvae} proposed a latent dynamics learning framework called the Newtonian Variational Autoencoder (NewtonianVAE).  Inspired by Newton's second law, a linear dynamic system in hidden space is defined, specifically by considering a rigid-body system with $D$ degrees of freedom and modeling the configuration of this system by a set of coordinates, $\bm{x}\in R^D$. Its dynamics are defined as
\begin{equation}
    \begin{split}
        &\frac{d\bm{x}}{dt}=\bm{v}\\
        &\frac{d\bm{v}}{dt} = A(\bm{x},\bm{v})\cdot \bm{x} + B(\bm{x},\bm{v})\cdot \bm{v} +C(\bm{x},\bm{v})\cdot \bm{u}
    \end{split}
\end{equation}
where $\bm{u}$ is the given actuation. To incorporate the above dynamic equations into a VAE,  positions $\bm{x}$ are treated as random variables with velocities $\bm{v}$ being intermediate variables. Specifically, velocities are computed as $\bm{v}_t=\frac{\bm{x}-\bm{x}_{t-1}}{\Delta t}$ with time interval $\Delta t$. The conditional distribution of $\bm{x}_t$ given $\bm{u}_t$ now becomes
\begin{equation}
    \begin{split}
        &p(\bm{x}_t|\bm{x}_{t-1},\bm{u}_{t-1}; \bm{v}_t) \sim \mathcal{N}(\bm{x}_{t-1}+\Delta t\cdot \bm{v}_t, \sigma^2)\\
        &\text{with}\quad \bm{v}_t = \bm{v}_{t-1} + \Delta t\cdot (A\bm{x}_{t-1}+B\bm{v}_{t-1}+C\bm{u}_{t-1})
    \end{split}
    \label{eq:newtonVAE}
\end{equation}
where $A$, $B$, and $C$ are estimated through a neural network, $f$, whose input consists of current system states (i.e., $\{\bm{x}_t, \bm{v}_t, \bm{u}_t$\}).  Newtonian VAE then outputs 2D images given the estimated $\bm{x}$.
This is in contrast with the existing approaches that simply assume a Gaussian distribution over $p(\bm{x}_t|\bm{x}_{t-1},\bm{u}_{t-1})$ without considering the underlying deterministic relationships among positions, forces, and velocities, as shown in Eq.~(\ref{eq:newtonVAE}).  Similarly, to incorporate the physics behind the transport dynamics governed by advection-diffusion PDEs, Liu et al.~\cite{liu2021discovering} proposed a learning framework based on an autoencoder with the advection-diffusion equation explicitly incorporated. Two hidden variable outputs from the encoder are physically meaningful, representing velocity field and diffusion field, respectively, within the advection-diffusion equation. To model lake temperature, Daw et al.~\cite{daw2020physics} introduced a physically meaningful intermediate variable for the proposed monotonicity-preserving long short term memory (LSTM) architecture. Specifically, the density value, as the intermediate variable of the LSTM, is enforced to monotonically increase as the depth increases, which is a crucial characteristic of lake temperature. A similar idea is applied in modeling drag forces acting on each particle in moving fluids~\cite{muralidhar2020phynet}. Muralidhar et al.~\cite{muralidhar2020phynet} proposed a PhyNet, where physics-constrained intermediate variables are introduced into a convolutional neural network (CNN) architecture. Specifically, two intermediate variables, characterizing velocity field and pressure field, respectively, are introduced into CNN for drag force prediction.

Besides introducing physically meaningful variables, another approach is to directly map some of the neural network parameters to physically meaningful parameters. These physics parameters can either be non-modifiable during training or be fine-tuned through learning from observations. In geophysics, neural networks have been considered for modeling the dynamic process of seismic waveform inversion~\cite{sun2020theory}.  To mimic seismic wave propagation, a theory-guided recurrent neural network (RNN) is proposed; an RNN is specially designed for solving the governing differential equations with some of the parameters assigned as the physics parameters in governing physics equations. In particular, given the wave equation discretized in the time domain, the wave field at the next time step (i.e., $u(t+\Delta t)$) is calculated in terms of two previous time steps (i.e., $u(t)$ and $u(t-\Delta t)$) as
\begin{equation}
\resizebox{.85\hsize}{!}{$\begin{split}
    u(r, t+\Delta t) &= v^2(t)\Delta t^2[\nabla^2u(r,t) \\&- s(r,t)\delta(r-r_s)]+2u(r,t)-u(r,t-\Delta t)
\end{split}$}
    \label{eq:wave-equation}
\end{equation}
where $\nabla^2$ is the spatial Laplacian operator, $r$ stands for position, and $s$ is the source function. The symbolic computation of $u(r, t+\Delta t)$ given $u(r,t)$ and $u(r,t-\Delta t)$ is directly implemented via a neural network, whose 
trainable parameters correspond to physics parameters $v^2\Delta t^2$ in the wave equation Eq.~(\ref{eq:wave-equation}).

\paragraph{Integrating through layers} 
The most representative type of knowledge that is usually integrated through neural network layers involves symmetries.  Symmetry usually refers to a set of invertible transformations $g$, such as translation, rotation, or scaling. 

Equivariance and invariance as representative types of symmetry have been widely considered. An invariant function is a mapping such that the output space is not affected by the symmetry transformations in the input space, and an equivariant function relaxes the invariant function. It states a mapping such that symmetries in the input space can be preserved in the output space. Mathematically, assume a symmetry transformation, $g$, and a function, $F$, mapping from $X$ to $Y$. $F$ is then said to be equivariant to $g$ if 
\begin{equation}
    F(g\cdot x) = g\cdot F(x)
\end{equation}
The symmetry transformation $g$ on the input space $X$ is preserved on the output space $Y$. $F$ is said to be invariant if 
\begin{equation}
    F(g \cdot x) = F(x)
\end{equation}
In other words, the output $y = F(x)$ is not affected by the symmetry transformation, $g$, acting upon the input space, $X$. Invariance can be a special case of equivariance, and vice versa. For example, transformation $g$ is an identify transformation with $g\cdot F(x) = F(x)$.  

\begin{figure}[ht!]
    \centering
    \includegraphics[width = 3in, height = 1.2in]{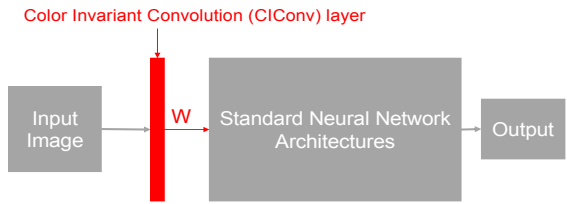}
    \caption{A physics layer (marked as red) for illumination invariant features~\cite{lengyel2021zero}. The new layer extracts illumination invariant features given input images and is customized based on Kubelka-Munk theory, termed as Color Invariant Convolution (CIConv) layer. }
    \label{fig:ch2-relatedworks-zeroshot}
\end{figure}

Equivariant or invariant neural network is designed to preserve symmetries. Illumination invariance features were explored~\cite{lengyel2021zero} in computer vision, where a knowledge-guided convolutional layer is incorporated into existing deep models. Consider a day-night domain adaptation problem, and illumination changes from source domain to target domain cause a distribution shift. To tackle the distribution shift issue, features that are invariant to illumination are desired, which can be derived from Kubelka-Munk theory. The Kubelka-Munk theory~\cite{kubelka1931article} models material reflections by defining the spectrum of light reflected from an object in the viewing direction. The calculation of illumination invariant features defined by Kubelka-Munk theory is directly implemented through the proposed Color Invariant Convolution (CIConv) layer, as shown in Figure~\ref{fig:ch2-relatedworks-zeroshot}. 
Similarly, in turbulence modeling, rotational invariance states that the physics of a fluid flow does not depend on the orientation of the coordinate of an observer and is a fundamental physical principle. To embed the rotational invariance into a neural network, Ling et al.~\cite{ling2016reynolds} defined a tensor basis neural network (TBNN), where the NN architecture is modified by adding a higher order multiplicative layer. Particularly, TBNN has one additional input layer accepting the tensor basis, and its last hidden layer performs a pair-wise multiplication using this tensor basis input layer to provide the output. The modified architecture ensures the prediction lie on a rotationally invariant tensor basis. By incorporating rotational invariance, TBNN achieves improved accuracy in predicting the normalized Reynolds stress anisotropy tensor. In the application to molecular dynamics, Anderson et al.~\cite{anderson2019cormorant} proposed a rotation invariant neural network, named Cormorant, whereby the behavior and properties of complex many-bodied physics systems are learned. Each neuron in Cormorant explicitly corresponds to a subset of atoms. Given specified neurons, the activation layer is ensured to be covariant to rotations, making the proposed Cormorant is guaranteed to be rotationally invariant. 


Equivariance has also been explored. Wang et al.~\cite{wang2020incorporating} showed that existing spatiotemporal deep models can achieve improved generalization ability by incorporating symmetries through equivariant functionality. More specifically, they consider four types of equivariance: time and space translation equivariance, rotation equivariance, uniform motion equivariance, and scale equivariance. These symmetries are incorporated into neural networks using customized equivariant
layers.  Through the composition of equivariant functions of layers, the network becomes equivariant.  

In the real physical world, however, symmetries could be brittle.  For example, a small perturbation can easily cause the discontinuous transition of a dynamic system or break the rotational symmetry of a pendulum system. Small perturbations often occur, and these can cause significant difference by accumulating over time. As a result, enforcing equivariance through customized layers as a hard inductive bias, could be problematic. To address this problem,  Finzi et al.~\cite{finzi2021residual} recently proposed a soft way to impose equivariance constraints, 
whereby the proposed neural network architecture consists of a hybrid of restrictive layers and flexible layers. Restrictive layers are strictly constrained while flexible layers are unconstrained. Through a mixture of two types of layers, the equivariance is introduced as a flexible inductive bias. 

\paragraph{Integrating through connectivity} 

Given physical dependencies among objects, connectivity among neural network neurons can be manually specified. To model dynamics with multiple objects involved, physically plausible interactions are employed to design NN connectivity.  Neural Physics Engine (NPE), a differentiable physics simulator that combines symbolic structure with gradient-based learning, has been proposed~\cite{chang2016compositional}. Different from conventional physics engines based on mechanistic models, NPE is realized as a learning-based neural network while remaining generalizable across different scenarios.
The development of NPE considers the fact that physics is Markovian, both temporally and spatially. Temporal Markovian allows the NPE to predict system states by only considering the states at the current step. Spatial Markovian allows the NPE to factorize the interaction dynamics into pairwise interactions. The NPE consists of a symbolic model structure mimicking pairwise interactions among objects, realized as a neural network. The NPE takes 2D observations as input and performs forward dynamics to predict the motions of objects in the future.  A very similar work was independently developed for n-body interaction systems~\cite{battaglia2016interaction}.

\subsubsection{Training-level Integration} 
One of the most common techniques to integrate scientific knowledge into deep models is through the training of deep models. In particular, constraints over the outputs of deep models are derived from scientific knowledge and are used as regularization terms for training a deep model. An augmented training objective is generally expressed as
\begin{equation}
    \label{eq:ch2-physics-aug-loss}
    \mathcal{L} = \mathcal{L}_{TRN}(y^{GT}, y^{Pred}) + \lambda \mathcal{L}_{PHY}(y^{Pred}; x)
\end{equation}
$\mathcal{L}_{TRN}(y^{GT}, y^{Pred})$ represent standard training loss given predictions $y^{Pred}$ and ground truth labels $y^{GT}$. For classification task, $\mathcal{L}_{TRN}(y^{GT}, y^{Pred})$ is usually defined based on the cross-entropy loss. The physics-based regularization term $\mathcal{L}_{PHY}(y^{Pred}; x)$ corresponds to physical constraints, with an adjustable importance coefficient, $\lambda$. When the physics-based constraints are independent of inputs $x$, the regularization term is reduced to $\mathcal{L}_{PHY}(y^{Pred})$. Through $\mathcal{L}_{PHY}(y^{Pred}; x)$, training is guided towards producing models with physically consistent outputs. The computation of physics-based regularization, $\mathcal{L}_{PHY}(y^{Pred}; x)$, does not require annotations of observations and hence allows unlabeled data to be included in training, reducing dependencies on data. Physics-based regularization, $\mathcal{L}_{PHY}(y^{Pred}; x)$, can also be employed directly for training deep models in a label-free manner. 

$\mathcal{L}_{PHY}(y^{Pred}; x)$ can be explicit or implicit and can be flexibly employed in different deep learning frameworks. Explicit regularization is directly defined over the output of deep models based on domain knowledge, while implicit regularization is induced by the physics-based models that are embedded into a deep learning pipeline. In the following, we introduce the knowledge-guided model regularization under two distinguishable deep model frameworks: discriminative deep models and generative deep models.

\paragraph{Regularization with discriminative deep models}  Model regularization with scientific knowledge is widely seen in discriminative deep models. In the context of climate modeling, constraints derived from the conservation laws that a physical system should satisfy are encoded as a regularization term. In particular, a NN, $f$ maps input $x$ to output $y$ (i.e., $y=f(x)$). Conservation constraints $\mathcal{C}$ are summarized as a linear system (i.e., $\mathcal{C} = \{C\begin{bmatrix}x\\y\end{bmatrix} = 0\}$), 
where $C$ is a given constraint matrix. These physical constraints are then encoded as a regularization term on the NN outputs:
\begin{equation}
    \mathcal{L}_{PHY}(x, y) = ||C\begin{bmatrix}x\\y\end{bmatrix}|| 
\end{equation}
Evaluation results show that by adding this physics-guided regularization, prediction performance is improved for emulating cloud processes~\cite{beucler2019achieving, beucler2021enforcing}. Similarly, Zhang et al.~\cite{zhang2018deeppotential} proposed parameterizing atomic energy for molecular dynamics through a NN, whose loss function considers the conservation of kinetic and potential energy. Regularization for a physics system with non-conservative forces has also been considered~\cite{seo2021controlling}. In a double pendulum system with friction, the total energy of a system is decreasing due to the existence of friction. Decreasing of energy is formulated as a constraint $E(\bm{x}_{t+1}) < E(\bm{x}_t)$, where $E(\bm{x}_t)$ and $E(\bm{x}_{t+1})$ denote the total energy of the system at current time step and a future time step, respectively. Considering a neural network which takes the state at current time step (i.e., $\bm{x}_{t}$) as input and outputs the estimated state at the next time step (i.e., $\bm{y} := \hat{\bm{x}}_{t+1}$), the constraint can be integrated into the neural network through the regularization term as 
\begin{equation}
    \mathcal{L}_{PHY}(\bm{x}_t, \bm{y}) = \mathtt{ReLU}(E(\bm{y})-E(\bm{x}_t))
    \label{eq:14}
\end{equation}
where $E(\bm{x}_t)$ and $E(\bm{y})$ computes the energy of the system at the current and next time step, respectively. According to the decreasing of energy, $E(\bm{y})$ is expected to be smaller than $E(\bm{x}_t)$, leading to the constraint above. Similarly, a set of common physical properties of a dynamic system is considered in~\cite{djeumou2021neural}, where each of these physical properties is represented as an equality or inequality constraint. These physics-informed constraints are then incorporated into deep models as regularization via an augmented Lagrangian method.

PDEs (as described in Eq.~(\ref{eq:pde})) have been widely considered as constraints and are integrated into deep models as regularization terms. 
A physics-informed neural network (PINN)~\cite{raissi2019physics} is proposed for solving PDEs by leveraging NNs.  A PINN learns solutions, $u$, by using both the observed data and PDEs whereby PDEs serve as inductive bias. Consider the viscous Burgers' equation as an example~\cite{karniadakis2021physics}:
\begin{equation}
    \frac{\partial u}{\partial t} + u\frac{\partial u}{\partial x} = v\frac{\partial^2 u}{\partial x^2}
\end{equation} 
In a PINN, a feed forward NN predicts a PDE solution, $\Tilde{u}$, by taking positions $x$ and time $t$ as input. The objective function consists of a data loss term and an PDE residual,
\begin{equation}
    \mathcal{L} = \mathcal{L}_{data} + \lambda \mathcal{L}_{PDE}
\end{equation}
where $\mathcal{L}_{data}$ measures the difference between predicted PDE solution $\Tilde{u}$ and given $u$ at certain positions and time step (i.e., $\mathcal{L}_{data} = \sum_{i=1}^{N_{data}}(\Tilde{u}(x_i, t_i) - u_i)^2$). $\mathcal{L}_{PDE}$ measures the PDE residual of predicted solution $\Tilde{u}$ at position and time step: 
\begin{equation}
    \mathcal{L}_{PDE} = \sum_{j=1}^{N_{PDE}}(\frac{\partial \Tilde{u}}{\partial t} + u\frac{\partial \Tilde{u}}{\partial x} - v\frac{\partial^2 \Tilde{u}}{\partial x^2})^2|_{(x_j, t_j)}
\end{equation}
Partial derivatives are calculated through numerical estimator given predicted solution $\Tilde{u}$.
Data points $\{(x_i, t_i)\}_{i=1}^{N_{data}}$ for data loss and $\{(x_j, t_j)\}_{j=1}^{N_{PDE}}$ for PDE residual are collected separately. 
$\lambda$ is the coefficient for the regularization term. In PINN, PDEs are directly encoded as a regularization term by measuring the solution residuals to constrain the model parameters. In the context of human body pose estimation, physics mechanics is employed to ensure physically plausible estimations, where the Euler-Lagrange equation represented as an ODE is derived, and is encoded as a soft constraint for model regularization.  
By integrating the Euler-Lagrange equation into data-driven deep models, the estimated 3D body pose is ensured to be physically plausible~\cite{li2019estimating}.

Physics-guided functions derived from domain knowledge have been used for training deep models in a label-free manner. Stewart and Ermon~\cite{stewart2017label} proposed a label-free supervision of NNs with physics equations.  The goal of this paper was to supervise NNs by specifying constraints $g$ that should hold over the output space  $f$ instead of using labels. The loss function then becomes
\begin{equation}
    \hat{f}^* = \arg\min_{f\in F} \sum_{i=1}^ng(x_i, f(x_i)) + R(f)
\end{equation}
where $R(f)$ refers to an additional regularization term penalizing model complexity. One example provided in the paper is to track an object performing free fall. The training of a regression network is formulated as a structured prediction problem operating on a sequence of $N$ images (i.e., $(\mathcal{R}^{H\times W\times 3})^N \rightarrow \mathcal{R}^N$).  Newton's second law of gravity (free fall motion) is represented as the algebraic equation and is directly incorporated into the loss function for training. 
In particular, for an object in free fall, its height at $i$-th time step with time interval $\Delta t$ is computed as $y_i = y_0 + v_0(i\Delta t) + a(i\Delta  t)^2$, with $y_0$ and $v_0$ being initial height and velocity. $a=-9.8m\backslash s^2$ is the fixed acceleration for an object performing free fall. Any predicted trajectory $f(\bm{x})$ hence should fit such parabola with fixed curvature. A loss is then defined measuring the fitting residual, 
\begin{equation}
\begin{split}
    g(\bm{x}, f(\textbf{x})) &= g(f(\bm{x})) = ||\hat{\bm{y}}-f(\bm{x})||_1\\
    &=||\textbf{a} + A(A^TA)^{-1}A^T(f(\textbf{x}) - \textbf{a}) - f(\bm{x})||_1
\end{split}
\end{equation}
where
\begin{equation}
    A = \begin{bmatrix}\triangle t & 1\\ 2\triangle t & 1 \\ 3\triangle t & 1\\
    ...\\N\triangle t & 1\end{bmatrix}
\end{equation}
and \textbf{a} = [$a\triangle t^2, a(2\triangle t)^2,...,a(N\triangle t)^2$] with $a=-9.8m\backslash s^2$. Besides, the algorithmic supervisions derived based on well-established algorithms have been used for training a neural network such that direct supervisions from ground truth annotations are no longer required~\cite{petersen2021learning}.

For all the related works discussed above, the relative importance between generic knowledge and data information is pre-defined by the design of training objective, and is not adjustable after training. However, the relative importance could vary with different inputs. For example, given an unseen input, data-based prediction is less reliable and knowledge is playing a more important role in the final prediction.
Seo et al.~\cite{seo2021controlling} proposed a framework where the relative importance of generic knowledge compared to data information is adjusted by a controlling parameter $\alpha$. The controlling parameter $\alpha$ is assumed to be a random variable following a pre-defined distribution $p(\alpha)$. Two sets of latent representations $\bm{z}_r$ and $\bm{z}_d$ extracted  from input data $\bm{x}$ correspond to generic knowledge and data information, respectively. The final latent features are obtained as $\bm{z} = \alpha \bm{z}_r \oplus (1-\alpha)\bm{z}_d$ and are used for generating final predictions $\bm{\hat{y}}$. Two sets of losses $\mathcal{L}_{rules}$ and $\mathcal{L}_{data}$ are defined based on generic knowledge and annotations from downstream tasks, respectively. $\mathcal{L}_{rules}$, being a function of inputs $\bm{x}$ and outputs $\bm{y}$, measures the violation of the rules derived based on prior knowledge specific to a target downstream task (e.g., Eq.~\ref{eq:14}). The final training loss is computed as the expected loss over $p(\alpha)$, i.e.,
\begin{equation}
    \mathcal{L} = \mathbb{E}_{\alpha\sim p(\alpha)}[\alpha\mathcal{L}_{rule} + \rho(1-\alpha)\mathcal{L}_{data}]
\end{equation}
where $\rho$ is a scale parameter to balance the units of two loss terms. Through the proposed framework, $\alpha$ denoting the relative importance becomes a variable during testing.

The physics-guided regularization terms discussed above are all explicitly defined over the output space of deep models. Physics-guided regularization terms can also be implicit (i.e., induced by a physics-based model embedded as one intermediate primitive of a neural network pipeline)~\cite{xie2020deep, wu2017learning, qiao2021differentiable}. For example, Wu et al.~\cite{wu2017learning} constructed a system for understanding physical scenes without human annotations. The core of the system is a physical world representation that is first recovered by a perception module and then utilized by a simulation engine. The perception module is a deep neural network that is self-supervised without annotation. The simulation engine, consisting of a physics engine and graphics engine, is aimed at generating physics predictions. 
\begin{figure}[ht!]
    \centering
    \includegraphics[width = 3in]{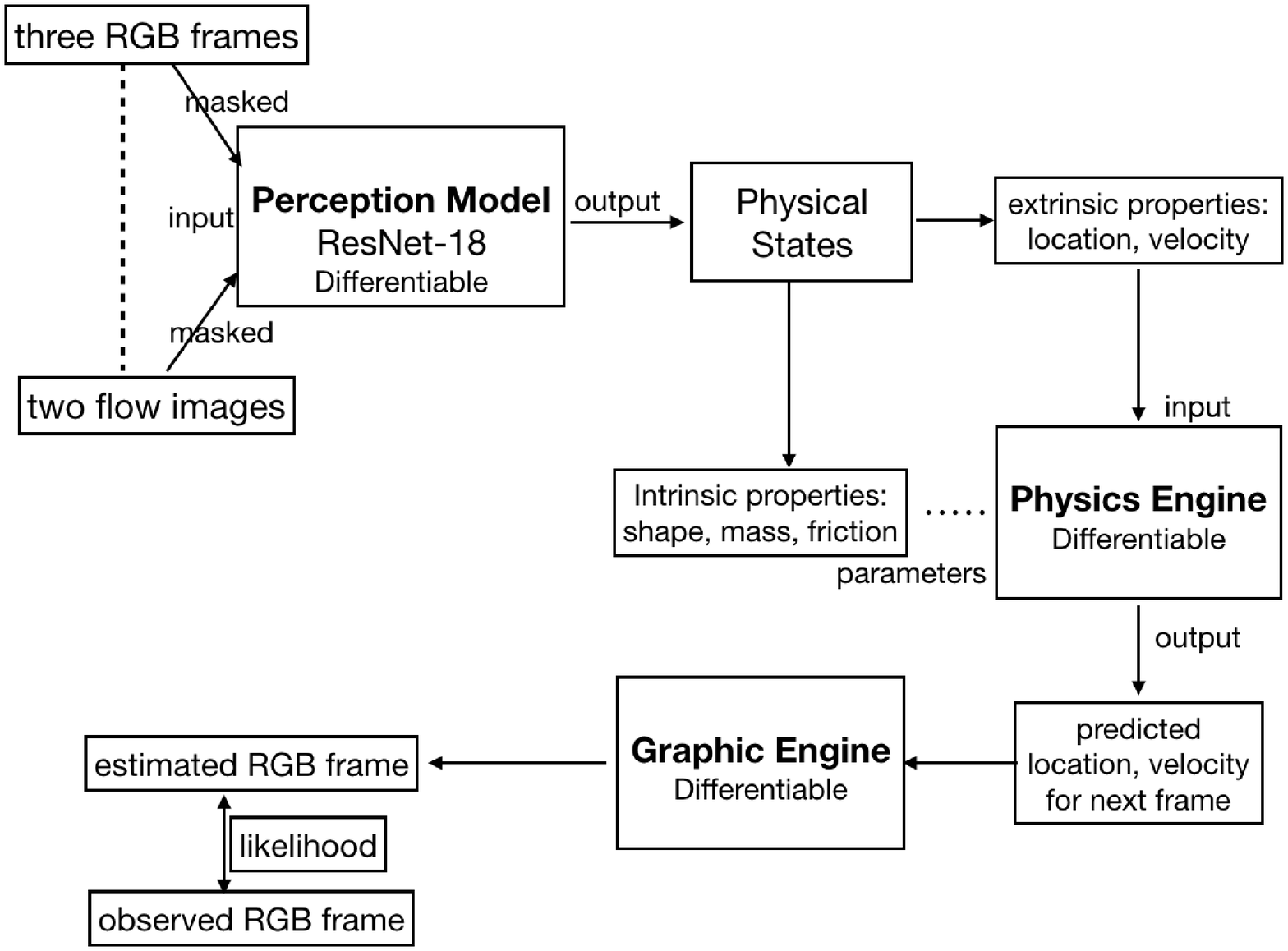}
    \caption{Physical scene understanding without human annotations~\cite{wu2017learning}. A physics engine and graphic engine are concatenated to a perception model. By aligning the rendered RGB images with observed ones, the perception model captures meaningful object features without annotations. The knowledge represented by the physics and graphic engines is encoded into the loss function implicitly. }
    \label{ch2-relatedworks-billiardtable}
\end{figure}
The proposed pipeline is evaluated on a synthetic billiard table experiment, as shown in Figure~\ref{ch2-relatedworks-billiardtable}. Given the features extracted by the perception model, the physics engine predicts the future physics states of a system by following motion equations, and the graphic engine renders RGB images given the predicted physics states. The perception model is learned by maximizing the likelihood of estimated RGB images given observed sequences.

\paragraph{Regularization with generative deep models} 
Besides discriminative models, knowledge has also been applied to deep generative models.
To generate realistic results, different generative models have explored the usage of domain knowledge. 
For example, 
to efficiently employ a generative adversarial network (GAN) for generating solutions of PDE-governed complex systems,  Wu et al.~\cite{wu2020enforcing} proposed a statistically constrained GAN, where a statistical regularization term is derived measuring the distance between covariance structures of training samples and generated samples, respectively, 
\begin{equation}
    \mathcal{L}_c(D,G) = \mathcal{L}(D,G) + \lambda d(\Sigma(p_{data}), \Sigma(p_{G}))
\end{equation}
where $D$ and $G$ represent standard discriminator and generator, respectively, and $L(D,G)$ represent the standard training loss for GAN. $\Sigma(p_{data})$ and $\Sigma(p_{G})$ represents the covariance structures of the distributions of training data and generated samples, respectively. $d(\cdot)$ represents a distance measurement, such as the Frobenius norm. $\lambda$ denotes the coefficient of the regularization term.  Introducing the statistical constraints on the lower dimensional manifold (i.e., covariance structure) helps reduce the searching space for finding desired solutions in high dimension. As a result,  not only is the amount of data reduced, but the training time needed to converge on solutions is shorten as well, as demonstrated in solving turbulence PDEs. 
In the application to computational materials science, Shah et al.~\cite{shah2019encoding} proposed a deep generative model named InvNet, through which synthetic structured samples satisfying desired physical properties are generated. InvNet is an extension of conventional GAN, whereby an additional invariance checker is introduced, along with a traditional generator and discriminator. The invariance checker is introduced as an intermediate primitive, based on which an implicit knowledge-guided regularization term is defined. An invariance loss is defined based on the proposed invariance checker, measuring the violation of invariance. Through the invariance loss, the generated samples are encouraged to satisfy certain invariances (such as motif invariance, planting a predefined motif in all synthetic images at a fixed location). Besides knowledge represented as PDEs, physical connectivity and stability have been considered for realistic 3D shape generation. Mezghanni et al.~\cite{mezghanni2021physically} proposed incorporating physical constraints into a deep generative model, whereby the physical constraints capture both the connectivity of 3D components and the physical stability of the 3D shape. Fully differentiable physical loss terms are then defined for integrating physics constraints into neural networks. 
Specifically, a neural stability predictor, implemented as a neural network classifier and pre-trained with simulated data, is proposed for enforcing physical stability constraints. For each synthetic 3D shape in the simulated data, its stability is labeled by the Bullet physics engine.  The stability constraint is encoded via the pre-trained stability classifier and is integrated into the deep model via the stability loss. 


Variational AutoEncoder (VAE) has also been explored with physics integrated for robust and interpretable generative modeling~\cite{yin2021augmenting, linial2021generative}. In particular, physics knowledge represented as PDEs is integrated into VAEs. Latent variables of VAEs are subject to the constraints defined by PDEs. Furthermore, in~\cite{takeishi2021physics}, instead of assuming a full access to the complete expression of PDEs, only part of the PDE is assumed to be known with latent variables of VAE being partially grounded with physics meaning for the known part of PDEs. The rest of unknown PDEs are modelled in a data-driven manner.


Leveraging probabilistic framework, uncertainty quantification has been considered in physics-informed deep models~\cite{yang2019adversarial, zhu2019physics, geneva2020modeling, karumuri2020simulator}.  
Zhu et al.~\cite{zhu2019physics} considered a physics-informed CNN solving PDEs with uncertainty quantification. The uncertainty is originated from the randomness of the physics parameters $\bm{\lambda}$, which is denoted as random vector $\bm{x}=\{\bm{\lambda}_1, ....,\bm{\lambda}_{n_s}\}$, where $n_s$ is the total number of possible physics parameter settings and can be very high. Correspondingly, the solution of PDE $u$ with respect to each possible physics parameter setting becomes $\bm{y}=\{u_1, u_2, ..., u_{n_s}\}$. The task is to model $p_{\theta}(\bm{y}|\bm{x})$ given a set of observations $\mathcal{D}_{input}=\{\bm{x}^{(i)}\}_{i=1}^N$ with $\bm{x}^{(i)}\sim p(\bm{x})$. $\bm{\theta}$ denote the neural network parameters to be learned. In the meantime, uncertainty of PDE solutions is modeled via the variance $Var[\bm{y}]$.  To train the NN, instead of using labeled data, an energy-based model is defined based solely on the PDE and its boundary conditions, from which we obtain the reference density.  In particular, the reference density $p_{\beta}(\bm{y}|\bm{x})$ follows a Boltzmann-Gibbs distribution:   
\begin{equation}
    p_\beta (\bm{y}|\bm{x}) = \frac{\exp (-\beta E(\bm{y, \bm{x}}))}{Z_\beta(\bm{x})}
\end{equation}
with $\beta$ being a learnable parameter of the constructed energy-based model. Energy function $E(\bm{y, \bm{x}})=V_{PDE}(\bm{y}, \bm{x})+\lambda V_{boundary}(\bm{y})$ measures the violation of the PDE and boundary conditions. $\lambda$ is a tunable hyper-parameter. Physics equations are encoded into the energy-based probabilistic model. In the end, the NN is trained by minimizing the KL divergence between the estimated distribution and the reference distribution: 
\begin{equation}
    \min_{\beta, \bm{\theta}} D_{KL}(p(\bm{x})p_{\bm{\theta}}(\bm{y}|\bm{x})||p(\bm{x})p_{\beta}(\bm{y}|\bm{x})) 
    \label{eq:physics-uncertainty-KL}
\end{equation}
Through the training process, physics equations are integrated into the NN. Along the same line, Yang and Perdikaris~\cite{yang2019adversarial} simplified the objective function by considering the lower bound of Eq.~(\ref{eq:physics-uncertainty-KL}). Following a similar idea, Karumuri et al.~\cite{karumuri2020simulator} employed a deep residual network (ResNet) for solving elliptic stochastic PDEs in a label-free manner. In particular, the physics-informed loss function is defined as the expectation of PDE residuals over the probability distribution of stochastic variables. All three works discussed above focus on time-independent physics systems without evolution over time. Geneva and Zabaras~\cite{geneva2020modeling} extended the idea to a dynamic system, where an auto regressive network is employed for predicting future physics states given a history of system states. 

\section{Deep Learning with Experiential Knowledge}
\label{sec:experiential_knowledge}
Besides scientific knowledge, experiential knowledge has been widely considered, as the major source of knowledge for neural-symbolic models. Experiential knowledge refers to well-known facts from everyday life, describing semantic properties of an object or semantic relationships among multiple objects. It is generally intuitive and is derived through long time observations or well-established studies. 
Unlike the scientific knowledge, experiential knowledge, though widely available, is descriptive and  imprecise.
Experiential knowledge containing semantic information can serve as a strong prior knowledge for predictive tasks in deep learning (e.g., regression or classification tasks), especially in the small-data regime where training data alone is insufficient in capturing relationships among variables~\cite{teshima2021incorporating}.

\subsection{Experiential Knowledge Identification}
Depending on the application domain, experiential knowledge may manifest in two types: entity properties and entity relationships.  
Entity relationships reveal semantic relationships among entities. They may be inferred from daily facts on relationships between named entities or be derived from well-established studies or theories, e.g., anatomy.
For example, human anatomy has been considered widely in computer vision for human body and facial behavior analysis~\cite{chen2013data, haker2009self, chen2021anatomy}.  For facial behavior analysis, facial anatomy knowledge may provide information on relationships among the facial muscles to produce natural facial expression.  Similarly, body anatomy may provide information on the relationships among the body joints to produce stable and physically plausible body pose and movement.  
The semantic relationships can be directly given or indirectly inferred from the existing ones. For example, from the facts that \texttt{Helen Mirren acted in the Ink Heart} and \texttt{Helen Mirren wins the Best Actress award}, one can infer that \texttt{The Ink Heart is nominated for an award}.  Inferred facts, however, are susceptible to errors.  Entity properties capture the knowledge about properties of entities. They can refer to ontological information describing hierarchical relations of concepts in the world perceived by humans~\cite{li2019large}. For example, \texttt{Rain is made of water}, and \texttt{Sea is a synonym for ocean}.  Linguistic knowledge, a major source of experiential knowledge, is analyzed in~\cite{yu2017visual, akula2021robust}. A large language model has been considered as inductive bias for an abstract textual reasoning task~\cite{rytting2021leveraging}. Linguistic knowledge such as textual explanations has been explored for language model refinement~\cite{yao2021refining}.

\subsection{Representation of Experiential Knowledge}
\begin{figure*}[ht!]
    \centering
    \includegraphics[width=6in, height=2.5in]{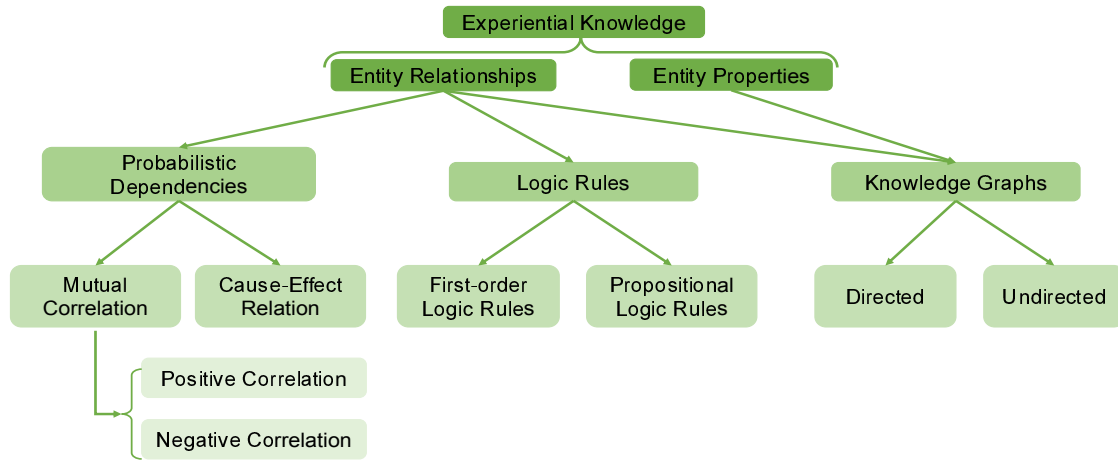}
    \caption{Taxonomy of experiential knowledge and its representations.} 
    \label{fig:taxonomy-experiential}
\end{figure*}
Representations of experiential knowledge vary with domains.  In general, the representations for experiential knowledge include probabilistic dependencies, logic rules and knowledge graphs. These representations capture both relationships and properties of entities revealed by the experiential knowledge. 
We illustrate the taxonomy of experiential knowledge and its representations in Figure~\ref{fig:taxonomy-experiential}.

\subsubsection{Probabilistic Dependencies} Due to inherent uncertainty, semantic relationships among objects are widely represented through probabilistic dependencies. States of objects are modelled in a probabilistic way, whereby relationships among objects are captured via probabilistic dependencies.  Relationships can be further divided into positive correlations and negative correlations.
Let's take facial action units (AUs) as an example. According to FACS~\cite{ekman1978facial},  AUs represent facial muscles and one facial muscle can control one or multiple AUs.  A binary AU can be on if the corresponding muscle is activated.
AU1 (inner brow raiser) and AU2 (outer brow raiser) usually occur together because they are controlled by the same muscle \textit{frontalis}. The pair AU15  (lip corner depressor) and AU24 (lip pressor) is another example of positive correlation, which is due to the fact that their underlying controlling muscles (i.e., \textit{depressor anguli oris} and \textit{orbicularis oris}, respectively) always move together. If two variables $X$ and $Y$ are positively correlated (e.g., $X$=``AU1'' and $Y$=``AU2'') with $X=\{$0, 1$\}$ and $Y=\{$0, 1$\}$, then we have
\begin{equation}
\begin{split}
        p(X=1,Y=1) &> p(X=1, Y=0)\\
        p(X=1,Y=1) &> p(X=0, Y=1)
\end{split}
\end{equation}
Considering negative correlation, AU12 (lip corner puller) and AU15 (lip corner depressor) cannot show up together as their corresponding muscles (i.e., \textit{Zgomaticus major} and \textit{depressor anguli oris}, respectively) are unlikely to be activated simultaneously. Negative correlation can be represented in a similar way. If two variables $X$ and $Y$ are negatively correlated (e.g., $X$=``AU12'' and $Y$=``AU15'') with $X=\{$0, 1$\}$ and $Y=\{$0, 1$\}$, then we have
\begin{equation}
\begin{split}
        p(X=1,Y=1) &< p(X=1, Y=0)\\
        p(X=1,Y=1) &< p(X=0, Y=1)
\end{split}
\end{equation}

\subsubsection{First-order and Propositional Logic} Logic can be categorized into first-order logic and propositional logic.
First-order logic (FOL)~\cite{enderton2001mathematical} employs logic rules to infer new experiential knowledge from existing experiential knowledge; it has been employed as an inference method to derive different types of knowledge, e.g., textual explanations~\cite{yao2021refining}. 
The formula of FOL is as follows: 
\begin{equation}
    R_1(\bm{x})\land R_2(\bm{x}) \land \cdots \land R_n(\bm{x}) \Rightarrow H(\bm{x})
\end{equation}
where $R_1(\bm{x}), R_2(\bm{x}), \cdots R_n(\bm{x})$ represent logic atoms. Each atom captures a known object property or relationship  through a predicate.  The atoms are combined through connective (e.g. conjunctive) operators to form the condition part of the logic rule.  
$H(\bm{x})$ is the implied result or conclusion of the logic rule.  It represents the new experiential knowledge derived from the logic rule.  The conditional part and the conclusion part of the rule are connected through the implication operator.   For example, we have
\begin{equation}
    \texttt{Smokes}(x) \Rightarrow \texttt{Cough} (x), \quad \forall x
\end{equation}
$\texttt{Smokes}(x)$ is an atom, with $\texttt{Smokes}$ being a predicate and $x$ as a logic variable. It captures the entity property knowledge that the person represented by variable $x$ smokes or not.  $\texttt{Cough}(x)$ is the implied result or derived knowledge, which captures the knowledge that the person coughs or not. 
The rule states that if the conditional part, $\texttt{Smokes}(x)$ is true, then  $\texttt{Cough}(x)$ is also true.

\subsubsection{Knowledge Graph} The knowledge graph is another symbolic representation of experiential knowledge,
that is used primarily to capture the semantic relationships among objects, whereby semantic knowledge is expressed in triple format: \texttt{(subject, predicate, object)}. The number of such triples is usually huge. In a knowledge graph, these triples are organized as a graph containing nodes and edges. Nodes represent subjects or objects, such as \texttt{animals} or \texttt{places}, as well as named entities, such as a person named \texttt{Mary Kelley}. Edges represent the predicate and connect pairs of nodes and describe the relationship between them. Besides, edges can also represent properties of an entity with nodes representing the attributes. Taking the triplet \texttt{(cat, attribute, paw)} as an example, nodes are \texttt{cat} and \texttt{claw}, and the relationship is \texttt{attribute}. This triple states a fact: “\texttt{The attribute of a cat is paw}".  Edges can be directed or undirected, for example, the food chain relationship between animals or the social relationship between people. Knowledge graphs can encode a large amount of commonsense, rules, and domain knowledge that capture semantic relationships and properties about entities. A knowledge graph hence is an important basic resource for obtaining experiential knowledge. For example, experiential knowledge of the semantic meanings of objects can be organized in a knowledge graph for image classification~\cite{Cui2019Image}. The Miscrosoft Concept Graph~\cite{ji2019microsoft} is another example of a knowledge graph, where
vertices in the Miscrosoft Concept Graph~\cite{ji2019microsoft} could represent food such as fruit, mammals such as dog and cat, or facilities such as bus and gas station. The edges indicate the relationships between concepts based on daily facts, such as \texttt{a cat is a mammal}, where \texttt{is} reflects relationship between \texttt{cat} and \texttt{mammal}.    

\subsection{Integration into Deep Models}
To integrate experiential knowledge into deep models, existing methodologies cover four types of integration: data-level, architecture-level, training-level, and decision-level as shown in Figure~\ref{fig:integrationlevels}. 
We introduce each group of the approaches in the following subsections. 

\subsubsection{Data-level Integration} 
Pseudo training data is usually considered to incorporate experiential knowledge represented as probabilistic dependencies, and is employed to augment existing training data. For example, Teshima and Sugiyama~\cite{teshima2021incorporating} proposed to incorporate the conditional independence relationships among variables 
into predictive modeling.  
A set of conditional independence relationships among variables are firstly extracted from prior knowledge. 
Training data is then augmented by generating synthetic data that satisfies the extracted conditional independence relationships.  Besides augmenting training data, pseudo data can also be employed for constructing a prior knowledge model. 
For facial AU recognition, Li et al.~\cite{li2013data} proposed to leverage pseudo data generated based on the knowledge for constructing a data-free prior model that captures the prior distribution of the target variables for downstream tasks. 
Constraints on parameters and variables are firstly derived from generic AU knowledge. Effective sampling methods are then proposed for generating the pseudo data satisfying the variable and parameter constraints.  A Bayesian network is then learned from pseudo data, serving as the data-free prior model. 
Similar idea has been exploited in upper body pose estimation task~\cite{chen2013data}. Four types of constraints (i.e., connectivity constraint, body length constraint, kinesiology constraint and symmetry constraint) are firstly derived from human anatomic knowledge. Synthetic data is then generated given these constraints, based on which a prior probabilistic model is learned. 

Besides, synthetic data can be generated from generic knowledge represented as Boolean rules~\cite{seo2021controlling}. 
For instance, the probability of $j$-th class (i.e., $\hat{\bm{y}}_j$) is higher when $k$-th input feature is bigger than a constant $a$ (i.e., $\bm{x}_k>a$). To incorporate this Boolean rule into a deep model, Seo et al.~\cite{seo2021controlling} proposed to augment each of the training data point $\bm{x}$ with a paired perturbed one $\bm{x}_{p}=\bm{x}+\delta \bm{x}$. $\delta \bm{x}$ is a small positive value for perturbation. The regularization is then be defined as
\begin{equation}
\resizebox{.87\hsize}{!}{$
    \mathcal{L}(\bm{x}, \bm{x}_p, \bm{\hat{y}}_j, \bm{\hat{y}}_{p,j}) = \texttt{ReLU}(\bm{\hat{y}}_j-\bm{\hat{y}}_{p,j})\cdot \mathbb{I}(\bm{x}_k<a)\cdot\mathbb{I}(\bm{x}_{p,k}>a)$}
\end{equation}
If $\bm{x}_k<a$ and $\bm{x}_{p,k}>a$, according to the prior knowledge, we should have $\bm{\hat{y}}_j - \bm{\hat{y}}_{p,j}<0$.  If the constraint is satisfied, we have $\mathcal{L}(\bm{x}, \bm{x}_p, \bm{\hat{y}}_j, \bm{\hat{y}}_{p,j})=0$. Otherwise, $\mathcal{L}(\bm{x}, \bm{x}_p, \bm{\hat{y}}_j, \bm{\hat{y}}_{p,j})>0$. 

\subsubsection{Architecture-level Integraion} 
Domain knowledge describing relationships between variables can be integrated into deep models through architecture design. We discuss the architecture-level integration methods for experiential knowledge represented as probabilistic dependencies, logic rules, and knowledge graphs, respectively.

\paragraph{Architecture design to incorporate probabilistic dependencies} 

One representative line of architecture-level integration is focused on the experiential knowledge represented as probabilistic dependencies, whereby a probabilistic model, constructed from the knowledge, serves as a prior model and is embedded as one layer of a neural network. Semantic relationships among variables can hence be incorporated into neural network in a probabilistic way. Usually, a probabilistic model is concatenated to the last layer of a neural network.  
A conditional random field (CRF) is usually employed, which takes hidden features from neural networks as input and outputs final predictions that satisfy the knowledge encoded in the CRF (e.g., among AUs~\cite{Corneanu2019}). In~\cite{chen2019deep}, 
a fully-connected CRF is concatenated to the last layer of a CNN to jointly perform facial landmark detection. By leveraging the fully-connected CNN-CRF, probabilistic predictions of facial landmark locations are obtained capturing the structural dependencies among landmark points.
For scene graph generation~\cite{suhail2021energy}, structured relationships among entities and relations are firstly captured via an energy-based probabilistic model. The energy-based probabilistic model takes the output of a typical scene graph generation model as input and refines it by minimizing the energy. 
Prior knowledge captured via a probabilistic model can also be leveraged to define the adjacency matrix of a graph convolutional network~\cite{song2021hybrid}. For facial action unit (AU) density estimation task, a Bayesian network is employed to capture the inherent dependencies among AUs. A probabilistic graph convolution is then proposed whereby its adjacency matrix is defined by the structure of the Bayesian network. In addition, a probabilistic model can also be introduced as a learnable intermediate layer of a neural network. A CausalVAE~\cite{yang2021causalvae} was proposed whereby a causal layer is introduced to the latent space of a variational auto-encoder (VAE). The causal layer essentially describes a structural causal model (SCM). Through the causal layer, independent exogenous factors are transformed to causal endogenous factors for causal representation learning.

\paragraph{Architecture design to incorporate logic rules}
Integration through the architecture of neural network is a conventional neural-symbolic approach to integrate symbolic logic rules into deep models. Logic rules are integrated into neural network architectures by introducing logic variables or parameters. Such an approach can be traced back to the 1990s, when knowledge-based artificial neural network (KBANN)~\cite{towell1994knowledge} and connectionist inductive learning and logic programming (CILP)~\cite{garcez1999connectionist} methods were introduced. 
More recently, logical neural network (LNN)~\cite{riegel2020logical} is proposed where each neuron represents an element in a logic formula, which can either be a concept (e.g., \texttt{cat}) or a logical connective (e.g., \texttt{AND}, \texttt{OR}). These works, however, are focused on leveraging neural networks for differentiable and scalable logic reasoning. 

Few works are proposed to improve deep models via customizing their architectures via logic rules. To leverage logic rules for improved deep model performance, logic rules are encoded into a Markov Logic Network (MLN), and the constructed MLN, serving as a prior model, is embedded into a neural network as an output layer for improved knowledge graph completion task~\cite{qu2019probabilistic}. Particularly, four types of logic rules are firstly identified for capturing the knowledge in a Knowledge graph: 
(1) composition rules: a predicate $R_k$ is composed of two predicates $R_i$ and $R_j$ if for any three variables $Y_1, Y_2, Y_3$ we have $R_i(Y_1, Y_2) \land R_j (Y_2, Y_2) \Rightarrow R_k (Y_1, Y_3)$; 
(2) Inverse rules: a predicate $R_i$ is an inverse of $R_j$ if for any two variables $Y_1$ and $Y_2$ we have $R_i(Y_1, Y_2) \Rightarrow R_j (Y_2, Y_1)$;
(3) Symmetric rules: a predicate $R$ is symmetric if for any two variables $Y_1$ and $Y_2$ we have $R(Y_1, Y_2) \Rightarrow R(Y_2, Y_1)$;
(4) Subset rules: a predicate $R_j$ is a subset of $R_i$ if for any two variables $Y_1$ and $Y_2$ we have $R_j(Y_1, Y_2) \Rightarrow R_i(Y_2, Y_1)$.
Given a set of identified logic rules $l\in\mathcal{L}$, a Markov logic network (MLN) defines a joint distribution of target variables $y$ of a triple as
\begin{equation}
    p(y) \propto \exp(\sum_{l\in\mathcal{L}}w_l\sum_{y}\phi_l(y_l))
\end{equation}
where $\phi_l$ is the potential function and is computed given the observed triples. Such an MLN is then concatenated to the last layer of a deep model whereby the confidence scores $w_l$ are learned via the deep model given the observed triples. By introducing the MLN, the task of predicting the missing triplets is re-formulated as inferring the posterior distribution of unseen configurations following the encoded logic rules.

\begin{figure}[ht!]
    \centering
\includegraphics[width=3.5in, height=1.2in]{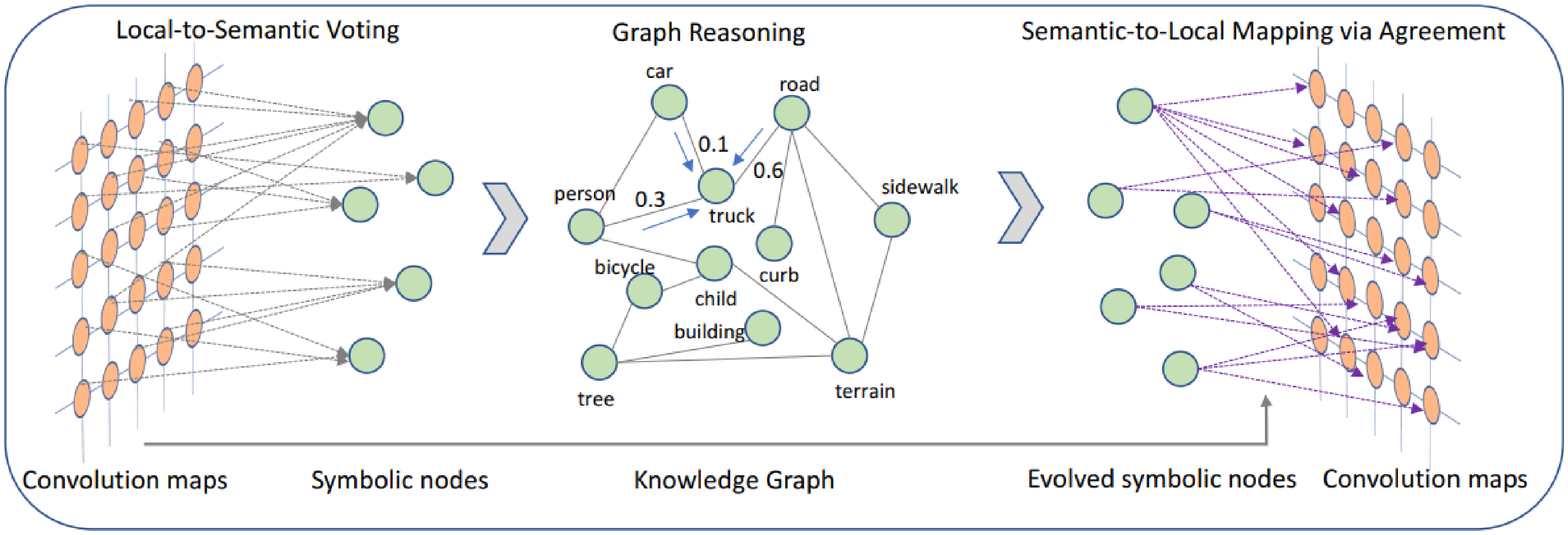}
    \caption{Image is adopted from~\cite{liang2018symbolic}. Nodes in the current convolutional layer are first mapped to symbolic nodes. Connectivities among the symbolic nodes are defined based on a knowledge graph. Through a graph reasoning based on the knowledge graph, evolved symbolic nodes are finally mapped to the nodes in the next convolutional layer.}
    \label{fig:SGR}
\end{figure}

\paragraph{Architecture design to incorporate knowledge graphs} The knowledge graph can also be integrated into the architecture of neural networks as one layer. 
Liang
et al.~\cite{liang2018symbolic} proposed a graph convolution with symbolic reasoning. Prior knowledge from the knowledge graph is specifically integrated into the neural network through a proposed symbolic graph reasoning (SGR) layer, as shown in Figure~\ref{fig:SGR}. 
To make the proposed SGR cooperate with a convolutional layer, local hidden features from the current convolutional layer are firstly transferred to features of corresponding symbolic nodes in SGR. Connectivities among the symbolic nodes in SGR are defined based on prior knowledge from the knowledge graph. Guided by prior knowledge, SGR then performs graph reasoning and updates features. Finally, the updated features are mapped onto local features in the next convolutional layer. 
To generate a medical report in an unsupervised manner in a cross visual domain and textual domain, a knowledge-driven encoder-decoder model is proposed to leverage knowledge graphs~\cite{liu2021auto}. The knowledge graph is encoded into the knowledge-driven attention module within the encoder. Taking an image and a medical report as inputs, the encoder first obtains the image embedding and the report embedding through standard deep models, respectively. An attention mechanism is then introduced whereby the embeddings are the queries and a knowledge graph is used to define the lookup matrix. The learned representations from the attention module bridge the vision and the textual domain by leveraging the knowledge graph. 
During training, the proposed model is learned by minimizing reconstruction error between generated and observed medical reports in the textual domain. 
During testing, the knowledge-driven encoder-decoder model can generate medical reports from medical images by leveraging experiential knowledge in knowledge graphs that is applicable to both visual and textual domains.
 
CRF has been leveraged to capture the experiential knowledge in a knowledge graph and is integrated as one layer of a neural network. Luo et al.~\cite{luo2020context} proposed a context-aware zero-shot recognition (CA-ZSL) method. Prior inter-object relation is extracted from knowledge graphs and encoded using a conditional random field (CRF). For an image containing $N$ objects, an image region and a class assignment of each object are denoted as $B_i$ and $c_i$, respectively, with $i=1,2,...,N$. The CRF model is then defined as
\begin{equation}
\begin{split}
        &P(c_1, ..., c_N | B_1, ..., B_N) \\
        &\propto \exp (\sum_{i}\theta(c_i | B_i) + \gamma \sum_{i |\neq j}\phi(c_i, c_j|B_i, B_j))
\end{split}
\end{equation}
where the unary potential $\theta(c_i|B_i)$ is estimated given extracted features of each object correspondingly. Pairwise potential $\phi(c_i, c_j|B_i, B_j)$ is estimated using both the extracted features and the knowledge graph. Semantic relations extracted from the knowledge graph are encoded in the pairwise potential functions, where $\gamma$ is a tunable hyper-parameter. Neural networks are trained by maximizing the log-likelihood. 
During testing, labels of unseen objects are inferred in a context-aware manner through a maximum-a-posteriori (MAP) inference in the learned CRF model. 

\subsubsection{Training-level Integration} 
Experiential knowledge is treated as prior bias that guides the training of a deep model. Constraints are obtained based on knowledge and are integrated as regularization terms into a deep model~\cite{xie2019embedding, donadello2017logic, diligenti2017semantic, li2019large}.
Regularization can be derived from probabilistic dependencies, logic rules, or knowledge graphs, and we discuss further in the paragraphs below.

\paragraph{Regularization with probabilistic dependencies} Probabilistic dependencies derived from knowledge 
about semantic relationships are commonly integrated into deep models through regularization.  Srinivas Kancheti et al.~\cite{srinivas2021matching} considered causal domain priors for regularizing neural networks during training, whereby the learned causal effects in NNs are enforced to match the prior knowledge on causal relationships through regularization. Consider a neural network $f$ with $d$ inputs and $C$ outputs, and for $j$th input, $\delta G^j$ is a $C\times d$ matrix containing prior causal knowledge (in terms of gradients). To enforce $f$ to be consistent with the prior knowledge, a regularization is defined
\begin{equation}
    R = \frac{1}{N}\sum_{j=1}^N \max\{0, ||\nabla_j f \odot M - \delta G^j||_1 - \epsilon\}
\end{equation}
where $M$ is a $C\times d$ binary matrix indicating the availability of prior knowledge and $N$ is the total number of training samples. $\nabla_j f$ is the $C\times d$ Jacobian of $f$ w.r.t. $j$th input. $\epsilon$ indicates an acceptable error margin, and $\odot$ is the element-wise product. Similarly, Rieger et al.~\cite{rieger2020interpretations} proposed to penalize model explanations that did not align with prior knowledge through an explanation loss.  For AU detection task, probabilistic relationships among AUs are derived from facial anatomic knowledge. Each of these probabilistic relationships is formulated as a constraint. In~\cite{zhang2018classifier}, a loss function measuring the satisfaction of each of these constraints is correspondingly defined and is employed for learning an AU detector. 
Differently, Cui et al.~\cite{cui2020knowledge} proposed to learn a Bayesian network (BN) to compactly capture a large set of constraints on AU relationships. 
The BN is then used to construct the expected cross-entropy loss to train a deep neural network for AU detection.

\paragraph{Regularization with logic rules} Logical knowledge is encoded as constraints for model regularization. Through regularization, a deep model is penalized if its output violates the constraints derived from logical rules. Xu et al.~\cite{xu2018semantic} proposed to combine the automatic reasoning technology of propositional logic with existing deep learning models. Propositional logic was encoded in the loss function through the proposed semantic loss. A sentence $\alpha$ in propositional logic is defined over variables $\bm{X}=\{X_1, ..., X_n\}$. The sentence is the semantic constraint  
to be imposed on the outputs of a neural network. Suppose that $p$ is a vector of probabilities, where each element $p_i$ denotes the predicted probability of variable $X_i$ and corresponds to a single output of the neural net. 
The semantic loss $L^s(\alpha, p)$ measures the violation of $\alpha$ given $p$ as
\begin{equation}
    L^s(\alpha, p) \propto -\log \sum_{\bm{x}\models \alpha}\prod_{i:\bm{x}\models X_i}p_i\prod_{i:\bm{x}\models \neg X_i}(1-p_i)
\end{equation}
$\bm{x}\models \alpha$ represents that the state $x$ satisfies the sentence $\alpha$. The larger the probability of the state satisfying the sentence, the smaller the semantic loss. The proposed semantic loss bridges neural network regularization with logic reasoning. It is effective for different applications, such as classification and preference ranking.

For relation prediction task,  a logic embedding network with semantic regularization (LENSR) was proposed~\cite{xie2019embedding}, where a propositional logic is integrated into a relation detection model. For a given image, the probability distribution of the relation predicate is first estimated using a standard visual relation detection model. Another probability distribution of the relation predicate is then proposed based on 
the propositional logic formula pre-defined given the input image.
Finally, a semantic regularization is defined to align these two probability distributions by minimizing their difference.
\paragraph{Regularization with knowledge graph}
The knowledge graph, a graphical representation of experiential knowledge, has also been employed for model regularization. Fang et al.~\cite{fang2017object} proposed to extract semantic consistency constraints from knowledge graph which are then used as regularization terms. Particularly, a consistency score between a pair of object $O$ and subject $S$ is calculated through random walk with restart
\begin{equation}
    \mathcal{C}_{O,S} = \lim_{T\rightarrow \infty} p(Y_T=Y_S|Y_0=Y_O; \alpha)
\end{equation}
where $\mathcal{C}\in R^{|O|\times |S|}$. $|O|$ and $|S|$ represent the total number of target objects and target subjects. $T$ is the total moving steps and $\alpha$ is the restart probability, meaning at each moving step, there is a probability of $\alpha$ to restart from the starting node instead of moving to one of the neighbors of node. The calculated matrix $\mathcal{C}$ is employed as constraints on semantic consistency and is used for regularizing the neural network for an object detection task. 
Similarly, Gu et al.~\cite{gu2019scene} proposed extracting external knowledge from KGs and applying image reconstruction to improve scene graph generation, especially when the dataset is biased, or annotations are noisy or missing. Object relationships are retrieved from ConceptNet as the external domain knowledge and are applied to refine object features through an object-to-image generation branch. The object-to-image generation branch reconstructs images based on object features and prior relational knowledge about the objects. Semantically meaningful object features can be learned by minimizing reconstruction error.

\subsubsection{Decision-level Integration} 
Predictions from deep models and from the prior knowledge can be directly combined through the joint top-down and bottom-up prediction strategy. Through integrating two sets of predictions, the final prediction can be more accurate and robust.

For open-domain knowledge-based visual question answering, Marino et al.~\cite{marino2021krisp} combined implicit and symbolic knowledge. Implicit knowledge refers to knowledge learned from data (e.g., raw texts). Symbolic knowledge refers to graph-based knowledge encoded in existing knowledge graphs (e.g., DBpedia~\cite{lehmann2015dbpedia} and ConceptNet~\cite{speer2017conceptnet}). The proposed KRISP model contains two sub-modules: implicit knowledge reasoning and explicit knowledge reasoning. Two sources of knowledge are then combined to generate the final output by using a late-fusion strategy. Through the late-fusion strategy, predictions from data and symbolic knowledge are directly combined, independent of the training of deep models. 

Two sets of predictions can be combined in a probabilistic way by following Bayes' rule. To obtain probabilistic knowledge-based predictions, a PGM model capturing prior knowledge is employed as a prior model and its prediction is obtained via a probabilistic inference. 
For AU recognition task, Li et al.~\cite{li2013data} considered a top-down and bottom-up integration, where a Bayesian network learned from generic knowledge serves as the top-down model and a data-driven model serves as the bottom-up model. Predictions from two models are then combined using Bayes' rule to produce the final predictions. Probabilistic knowledge-based predictions can also be directly defined based on the knowledge. For a knowledge graph completion task, Cui et al.~\cite{cui2021type} derived a prior distribution on relations based on type information. The prior distribution is then combined with bottom-up predictions from existing embedding-based models through Bayes' rule for final prediction.

\section{Discussion and Future Direction}
\begin{figure}[ht!]
    \centering
    \includegraphics[width=3.4in]{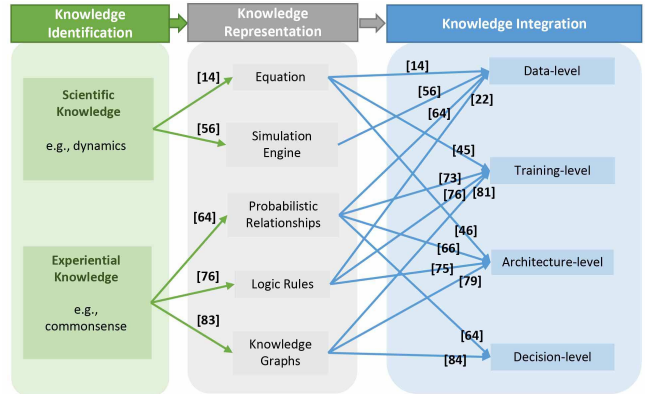}
    \caption{Prescriptive tree on knowledge-augmented deep learning}
    \label{fig:tree}
\end{figure}

In this survey, we reviewed traditional and prevailing technologies for knowledge-augmented deep learning, including knowledge identification, knowledge representation, and integration. We divided knowledge into two categories: scientific knowledge and experiential knowledge. Within each category, we introduced knowledge identification, representations, and integration into deep learning. As we have discussed, much work has been done to improve deep learning with related prior knowledge to yield data-efficient, generalizable, and interpretable deep learning models. To help readers better understand and apply the KADL to their work, we provide a prescriptive tree based on the summary of existing works (Table~\ref{tab:overall-summary}). The prescriptive tree as shown in Figure~\ref{fig:tree} serves as a recipe that includes different pathways to incorporate a particular type of prior knowledge into deep models. Each pathway consists of a specific  knowledge type, a knowledge representation format, and a knowledge integration method, accompanied with the references to related works. 

Despite these efforts, existing methods suffer from several shortcomings. In the following paragraphs, we discuss existing techniques and highlight promising directions to pursue in the future.

\paragraph{Diverse types of knowledge} 
Existing knowledge-augmented deep models explore different types of domain knowledge, including scientific knowledge and experiential knowledge. However, most of the knowledge explored are scientific knowledge in physics and symbolically represented experiential knowledge. 
Injecting well-established algorithmic knowledge into deep models has already began to attract researchers' attention whereby deep models are trained through algorithmic supervisions instead of ground truth annotations~\cite{petersen2021learning}. Furthermore, 
existing approaches are usually limited to one specific type of knowledge. For a certain application task, both scientific knowledge and experiential knowledge could exists from multiple sources. Hence, diverse types of knowledge can be combined jointly for improved deep model performance.

\paragraph{Effective knowledge integration}
Existing integration methods utilize synthetic data, model architecture design, regularization function, or prediction refinement. Among them, the majority of the integration methods are performed during training. Existing integration techniques hence heavily depend on the specific training procedure, through which
deep models are trained by considering two sources of information jointly, without 
explicitly differentiating data from knowledge. The problem can be addressed through decision-level fusion.  Decision-level knowledge integration scheme using a prior model to capture the domain knowledge attracts relatively less attention. Knowledge integration with a prior model to combine top-down prediction from the knowledge and bottom-up estimation from the data could be beneficial from several aspects. Firstly, the construction of a prior model is independent of deep models whereby deep models are initialized through observable data. Since the prior model and the deep model are built independent of the integration process, top-down and bottom-up integration process can be flexibly applied to any deep model and prior model. Secondly, knowledge integration is performed by following Bayes' rule in a principled manner. Refinement of data-based prediction based on knowledge becomes tractable and interpretable. 



\paragraph{Hybrid integration methods}
Existing methods tend to integrate scientific knowledge and experiential knowledge separately. In addition, they tend to employ one specific approach to perform the knowledge integration.
For certain application domains, both types of knowledge may exist at the same time. Hence, they should be integrated jointly to further improve the performance of the deep models. In addition, users always need to choose an integration approach. There is no universal integration scheme that applies to all types of knowledge, and it remains an open question on how to automatically integrate knowledge with data in an optimal way. Hence,
given the complementary nature of different integration methods, it may be beneficial to simultaneously employ different integration methods to exploit their 
respective strengths.



\paragraph{Knowledge integration with uncertainty} 
Existing works have explored the encoding of experiential knowledge in a probabilistic way such as capturing the uncertain relationships using a probabilistic graphical model. 
But in general, existing knowledge integration methods are deterministic methods, ignoring the underlying knowledge uncertainty and their impacts on deep model learning and inference.
Uncertainty not only exists with experiential knowledge but also with scientific knowledge.
For example, in physics, uncertainty arises from random physics parameters or unknown physics parameters or incomplete observations. Existing works along this line aim to measure the quality of PDE solutions governing a physics system and thus are subject to specific domain assumptions. Probabilistic tools, such as probabilistic graphical models (PGMs), are powerful in capturing uncertainties for experiential knowledge. However, few works explored the usage of PGMs in modeling uncertainties in scientific laws~\cite{enright2013bayesian,evers2011constructing}. It remains an open question to deep learning community on how to effectively and systematically model uncertainties in scientific knowledge for real-world application tasks.

\bibliographystyle{IEEEtran}
\bibliography{reference}

\begin{thebibliography}{100}
\providecommand{\url}[1]{#1}
\csname url@samestyle\endcsname
\providecommand{\newblock}{\relax}
\providecommand{\bibinfo}[2]{#2}
\providecommand{\BIBentrySTDinterwordspacing}{\spaceskip=0pt\relax}
\providecommand{\BIBentryALTinterwordstretchfactor}{4}
\providecommand{\BIBentryALTinterwordspacing}{\spaceskip=\fontdimen2\font plus
\BIBentryALTinterwordstretchfactor\fontdimen3\font minus
  \fontdimen4\font\relax}
\providecommand{\BIBforeignlanguage}[2]{{%
\expandafter\ifx\csname l@#1\endcsname\relax
\typeout{** WARNING: IEEEtran.bst: No hyphenation pattern has been}%
\typeout{** loaded for the language `#1'. Using the pattern for}%
\typeout{** the default language instead.}%
\else
\language=\csname l@#1\endcsname
\fi
#2}}
\providecommand{\BIBdecl}{\relax}
\BIBdecl

\bibitem{marcus2018deep}
G.~Marcus, ``Deep learning: A critical appraisal,'' \emph{arXiv preprint
  arXiv:1801.00631}, 2018.

\bibitem{besold2017neural}
T.~R. Besold, A.~d. Garcez, S.~Bader, H.~Bowman, P.~Domingos, P.~Hitzler, K.-U.
  K{\"u}hnberger, L.~C. Lamb, D.~Lowd, P.~M.~V. Lima \emph{et~al.},
  ``Neural-symbolic learning and reasoning: A survey and interpretation,''
  \emph{arXiv preprint arXiv:1711.03902}, 2017.

\bibitem{yu2021survey}
D.~Yu, B.~Yang, D.~Liu, and H.~Wang, ``A survey on neural-symbolic systems,''
  \emph{arXiv preprint arXiv:2111.08164}, 2021.

\bibitem{willard2020integrating}
J.~Willard, X.~Jia, S.~Xu, M.~Steinbach, and V.~Kumar, ``Integrating
  physics-based modeling with machine learning: A survey,'' \emph{arXiv
  preprint arXiv:2003.04919}, vol.~1, no.~1, pp. 1--34, 2020.

\bibitem{han2020integrating}
J.~Han, L.~Zhang \emph{et~al.}, ``Integrating machine learning with
  physics-based modeling,'' \emph{arXiv preprint arXiv:2006.02619}, 2020.

\bibitem{rai2020driven}
R.~Rai and C.~K. Sahu, ``Driven by data or derived through physics? a review of
  hybrid physics guided machine learning techniques with cyber-physical system
  (cps) focus,'' \emph{IEEE Access}, vol.~8, pp. 71\,050--71\,073, 2020.

\bibitem{rath2020boosting}
M.~Rath and A.~P. Condurache, ``Boosting deep neural networks with geometrical
  prior knowledge: A survey,'' \emph{arXiv preprint arXiv:2006.16867}, 2020.

\bibitem{nowack2018using}
P.~Nowack, P.~Braesicke, J.~Haigh, N.~L. Abraham, J.~Pyle, and A.~Voulgarakis,
  ``Using machine learning to build temperature-based ozone parameterizations
  for climate sensitivity simulations,'' \emph{Environmental Research Letters},
  vol.~13, no.~10, p. 104016, 2018.

\bibitem{deng2020integrating}
C.~Deng, X.~Ji, C.~Rainey, J.~Zhang, and W.~Lu, ``Integrating machine learning
  with human knowledge,'' \emph{Iscience}, vol.~23, no.~11, p. 101656, 2020.

\bibitem{von2019informed}
L.~von Rueden, S.~Mayer, K.~Beckh, B.~Georgiev, S.~Giesselbach, R.~Heese,
  B.~Kirsch, J.~Pfrommer, A.~Pick, R.~Ramamurthy \emph{et~al.}, ``Informed
  machine learning--a taxonomy and survey of integrating knowledge into
  learning systems,'' \emph{arXiv preprint arXiv:1903.12394}, 2019.

\bibitem{kim2021knowledge}
S.~W. Kim, I.~Kim, J.~Lee, and S.~Lee, ``Knowledge integration into deep
  learning in dynamical systems: An overview and taxonomy,'' \emph{Journal of
  Mechanical Science and Technology}, vol.~35, no.~4, pp. 1331--1342, 2021.

\bibitem{von2020combining}
L.~von Rueden, S.~Mayer, R.~Sifa, C.~Bauckhage, and J.~Garcke, ``Combining
  machine learning and simulation to a hybrid modelling approach: Current and
  future directions,'' in \emph{International Symposium on Intelligent Data
  Analysis}.\hskip 1em plus 0.5em minus 0.4em\relax Springer, 2020, pp.
  548--560.

\bibitem{sagel2020knowledge}
A.~Sagel, A.~Sahu, S.~Matthes, H.~Pfeifer, T.~Qiu, H.~Rue{\ss}, H.~Shen, and
  J.~W{\"o}rmann, ``Knowledge as invariance--history and perspectives of
  knowledge-augmented machine learning,'' \emph{arXiv preprint
  arXiv:2012.11406}, 2020.

\bibitem{chen2013data}
J.~Chen, S.~Nie, and Q.~Ji, ``Data-free prior model for upper body pose
  estimation and tracking,'' \emph{IEEE Transactions on Image Processing},
  vol.~22, no.~12, pp. 4627--4639, 2013.

\bibitem{liu2021discovering}
P.~Liu, L.~Tian, Y.~Zhang, S.~Aylward, Y.~Lee, and M.~Niethammer, ``Discovering
  hidden physics behind transport dynamics,'' in \emph{Proceedings of the
  IEEE/CVF Conference on Computer Vision and Pattern Recognition}, 2021, pp.
  10\,082--10\,092.

\bibitem{jaques2021newtonianvae}
M.~Jaques, M.~Burke, and T.~M. Hospedales, ``Newtonianvae: Proportional control
  and goal identification from pixels via physical latent spaces,'' in
  \emph{Proceedings of the IEEE/CVF Conference on Computer Vision and Pattern
  Recognition}, 2021, pp. 4454--4463.

\bibitem{daw2020physics}
A.~Daw, R.~Q. Thomas, C.~C. Carey, J.~S. Read, A.~P. Appling, and A.~Karpatne,
  ``Physics-guided architecture (pga) of neural networks for quantifying
  uncertainty in lake temperature modeling,'' in \emph{Proceedings of the 2020
  siam international conference on data mining}.\hskip 1em plus 0.5em minus
  0.4em\relax SIAM, 2020, pp. 532--540.

\bibitem{muralidhar2020phynet}
N.~Muralidhar, J.~Bu, Z.~Cao, L.~He, N.~Ramakrishnan, D.~Tafti, and
  A.~Karpatne, ``Phynet: Physics guided neural networks for particle drag force
  prediction in assembly,'' in \emph{Proceedings of the 2020 SIAM International
  Conference on Data Mining}.\hskip 1em plus 0.5em minus 0.4em\relax SIAM,
  2020, pp. 559--567.

\bibitem{beucler2019achieving}
T.~Beucler, S.~Rasp, M.~Pritchard, and P.~Gentine, ``Achieving conservation of
  energy in neural network emulators for climate modeling,'' \emph{arXiv
  preprint arXiv:1906.06622}, 2019.

\bibitem{beucler2021enforcing}
T.~Beucler, M.~Pritchard, S.~Rasp, J.~Ott, P.~Baldi, and P.~Gentine,
  ``Enforcing analytic constraints in neural networks emulating physical
  systems,'' \emph{Physical Review Letters}, vol. 126, no.~9, p. 098302, 2021.

\bibitem{zhang2018deeppotential}
L.~Zhang, J.~Han, H.~Wang, R.~Car, and E.~Weinan, ``Deep potential molecular
  dynamics: a scalable model with the accuracy of quantum mechanics,''
  \emph{Physical review letters}, vol. 120, no.~14, p. 143001, 2018.

\bibitem{seo2021controlling}
S.~Seo, S.~O. Arik, J.~Yoon, X.~Zhang, K.~Sohn, and T.~Pfister, ``Controlling
  neural networks with rule representations,'' \emph{arXiv preprint
  arXiv:2106.07804}, 2021.

\bibitem{sun2020theory}
J.~Sun, Z.~Niu, K.~A. Innanen, J.~Li, and D.~O. Trad, ``A theory-guided
  deep-learning formulation and optimization of seismic waveform inversion,''
  \emph{Geophysics}, vol.~85, no.~2, pp. R87--R99, 2020.

\bibitem{toth2019hamiltonian}
P.~Toth, D.~J. Rezende, A.~Jaegle, S.~Racani{\`e}re, A.~Botev, and I.~Higgins,
  ``Hamiltonian generative networks,'' \emph{arXiv preprint arXiv:1909.13789},
  2019.

\bibitem{choudhary2020physics}
A.~Choudhary, J.~F. Lindner, E.~G. Holliday, S.~T. Miller, S.~Sinha, and W.~L.
  Ditto, ``Physics-enhanced neural networks learn order and chaos,''
  \emph{Physical Review E}, vol. 101, no.~6, p. 062207, 2020.

\bibitem{djeumou2021neural}
F.~Djeumou, C.~Neary, E.~Goubault, S.~Putot, and U.~Topcu, ``Neural networks
  with physics-informed architectures and constraints for dynamical systems
  modeling,'' \emph{arXiv preprint arXiv:2109.06407}, 2021.

\bibitem{raissi2019physics}
M.~Raissi, P.~Perdikaris, and G.~E. Karniadakis, ``Physics-informed neural
  networks: A deep learning framework for solving forward and inverse problems
  involving nonlinear partial differential equations,'' \emph{Journal of
  Computational Physics}, vol. 378, pp. 686--707, 2019.

\bibitem{karniadakis2021physics}
G.~E. Karniadakis, I.~G. Kevrekidis, L.~Lu, P.~Perdikaris, S.~Wang, and
  L.~Yang, ``Physics-informed machine learning,'' \emph{Nature Reviews
  Physics}, vol.~3, no.~6, pp. 422--440, 2021.

\bibitem{li2019estimating}
Z.~Li, J.~Sedlar, J.~Carpentier, I.~Laptev, N.~Mansard, and J.~Sivic,
  ``Estimating 3d motion and forces of person-object interactions from
  monocular video,'' in \emph{Proceedings of the IEEE/CVF Conference on
  Computer Vision and Pattern Recognition}, 2019, pp. 8640--8649.

\bibitem{greydanus2019hamiltonian}
S.~J. Greydanus, M.~Dzumba, and J.~Yosinski, ``Hamiltonian neural networks,''
  2019.

\bibitem{zhong2019symplectic}
Y.~D. Zhong, B.~Dey, and A.~Chakraborty, ``Symplectic ode-net: Learning
  hamiltonian dynamics with control,'' \emph{arXiv preprint arXiv:1909.12077},
  2019.

\bibitem{cranmer2020lagrangian}
M.~Cranmer, S.~Greydanus, S.~Hoyer, P.~Battaglia, D.~Spergel, and S.~Ho,
  ``Lagrangian neural networks,'' \emph{arXiv preprint arXiv:2003.04630}, 2020.

\bibitem{allen2020lagnetvip}
C.~Allen-Blanchette, S.~Veer, A.~Majumdar, and N.~E. Leonard, ``Lagnetvip: A
  lagrangian neural network for video prediction,'' \emph{arXiv preprint
  arXiv:2010.12932}, 2020.

\bibitem{stewart2017label}
R.~Stewart and S.~Ermon, ``Label-free supervision of neural networks with
  physics and domain knowledge,'' in \emph{Thirty-First AAAI Conference on
  Artificial Intelligence}, 2017.

\bibitem{xie2020deep}
F.~Xie, A.~Chowdhury, M.~Kaluza, L.~Zhao, L.~L. Wong, and R.~Yu, ``Deep
  imitation learning for bimanual robotic manipulation,'' \emph{arXiv preprint
  arXiv:2010.05134}, 2020.

\bibitem{wu2017learning}
J.~Wu, E.~Lu, P.~Kohli, B.~Freeman, and J.~Tenenbaum, ``Learning to see physics
  via visual de-animation,'' \emph{Advances in Neural Information Processing
  Systems}, vol.~30, pp. 153--164, 2017.

\bibitem{qiao2021differentiable}
Y.-L. Qiao, J.~Liang, V.~Koltun, and M.~Lin, ``Differentiable simulation of
  soft multi-body systems,'' in \emph{Thirty-Fifth Conference on Neural
  Information Processing Systems}, 2021.

\bibitem{botev2021priors}
A.~Botev, A.~Jaegle, P.~Wirnsberger, D.~Hennes, and I.~Higgins, ``Which priors
  matter? benchmarking models for learning latent dynamics,'' 2021.

\bibitem{wang2020incorporating}
R.~Wang, R.~Walters, and R.~Yu, ``Incorporating symmetry into deep dynamics
  models for improved generalization,'' \emph{arXiv preprint arXiv:2002.03061},
  2020.

\bibitem{finzi2021residual}
M.~Finzi, G.~Benton, and A.~G. Wilson, ``Residual pathway priors for soft
  equivariance constraints,'' \emph{Advances in Neural Information Processing
  Systems}, vol.~34, 2021.

\bibitem{ling2016reynolds}
J.~Ling, A.~Kurzawski, and J.~Templeton, ``Reynolds averaged turbulence
  modelling using deep neural networks with embedded invariance,''
  \emph{Journal of Fluid Mechanics}, vol. 807, pp. 155--166, 2016.

\bibitem{mezghanni2021physically}
M.~Mezghanni, M.~Boulkenafed, A.~Lieutier, and M.~Ovsjanikov,
  ``Physically-aware generative network for 3d shape modeling,'' in
  \emph{Proceedings of the IEEE/CVF Conference on Computer Vision and Pattern
  Recognition}, 2021, pp. 9330--9341.

\bibitem{shah2019encoding}
V.~Shah, A.~Joshi, S.~Ghosal, B.~Pokuri, S.~Sarkar, B.~Ganapathysubramanian,
  and C.~Hegde, ``Encoding invariances in deep generative models,'' \emph{arXiv
  preprint arXiv:1906.01626}, 2019.

\bibitem{wu2020enforcing}
J.-L. Wu, K.~Kashinath, A.~Albert, D.~Chirila, H.~Xiao \emph{et~al.},
  ``Enforcing statistical constraints in generative adversarial networks for
  modeling chaotic dynamical systems,'' \emph{Journal of Computational
  Physics}, vol. 406, p. 109209, 2020.

\bibitem{geneva2020modeling}
N.~Geneva and N.~Zabaras, ``Modeling the dynamics of pde systems with
  physics-constrained deep auto-regressive networks,'' \emph{Journal of
  Computational Physics}, vol. 403, p. 109056, 2020.

\bibitem{anderson2019cormorant}
B.~Anderson, T.-S. Hy, and R.~Kondor, ``Cormorant: Covariant molecular neural
  networks,'' \emph{arXiv preprint arXiv:1906.04015}, 2019.

\bibitem{lengyel2021zero}
A.~Lengyel, S.~Garg, M.~Milford, and J.~C. van Gemert, ``Zero-shot day-night
  domain adaptation with a physics prior,'' in \emph{Proceedings of the
  IEEE/CVF International Conference on Computer Vision}, 2021, pp. 4399--4409.

\bibitem{chang2016compositional}
M.~B. Chang, T.~Ullman, A.~Torralba, and J.~B. Tenenbaum, ``A compositional
  object-based approach to learning physical dynamics,'' \emph{arXiv preprint
  arXiv:1612.00341}, 2016.

\bibitem{battaglia2016interaction}
P.~W. Battaglia, R.~Pascanu, M.~Lai, D.~Rezende, and K.~Kavukcuoglu,
  ``Interaction networks for learning about objects, relations and physics,''
  \emph{arXiv preprint arXiv:1612.00222}, 2016.

\bibitem{takeishi2021physics}
N.~Takeishi and A.~Kalousis, ``Physics-integrated variational autoencoders for
  robust and interpretable generative modeling,'' \emph{arXiv preprint
  arXiv:2102.13156}, 2021.

\bibitem{yin2021augmenting}
Y.~Yin, V.~Le~Guen, J.~Dona, E.~de~B{\'e}zenac, I.~Ayed, N.~Thome, and
  P.~Gallinari, ``Augmenting physical models with deep networks for complex
  dynamics forecasting,'' \emph{Journal of Statistical Mechanics: Theory and
  Experiment}, 2021.

\bibitem{linial2021generative}
O.~Linial, N.~Ravid, D.~Eytan, and U.~Shalit, ``Generative ode modeling with
  known unknowns,'' in \emph{Proceedings of the Conference on Health,
  Inference, and Learning}, 2021, pp. 79--94.

\bibitem{yang2019adversarial}
Y.~Yang and P.~Perdikaris, ``Adversarial uncertainty quantification in
  physics-informed neural networks,'' \emph{Journal of Computational Physics},
  vol. 394, pp. 136--152, 2019.

\bibitem{zhu2019physics}
Y.~Zhu, N.~Zabaras, P.-S. Koutsourelakis, and P.~Perdikaris,
  ``Physics-constrained deep learning for high-dimensional surrogate modeling
  and uncertainty quantification without labeled data,'' \emph{Journal of
  Computational Physics}, vol. 394, pp. 56--81, 2019.

\bibitem{karumuri2020simulator}
S.~Karumuri, R.~Tripathy, I.~Bilionis, and J.~Panchal, ``Simulator-free
  solution of high-dimensional stochastic elliptic partial differential
  equations using deep neural networks,'' \emph{Journal of Computational
  Physics}, vol. 404, p. 109120, 2020.

\bibitem{jia2019physics}
X.~Jia, J.~Willard, A.~Karpatne, J.~Read, J.~Zwart, M.~Steinbach, and V.~Kumar,
  ``Physics guided rnns for modeling dynamical systems: A case study in
  simulating lake temperature profiles,'' in \emph{Proceedings of the 2019 SIAM
  International Conference on Data Mining}.\hskip 1em plus 0.5em minus
  0.4em\relax SIAM, 2019, pp. 558--566.

\bibitem{jia2021physics}
X.~Jia, J.~Willard, A.~Karpatne, J.~S. Read, J.~A. Zwart, M.~Steinbach, and
  V.~Kumar, ``Physics-guided machine learning for scientific discovery: An
  application in simulating lake temperature profiles,'' \emph{ACM/IMS
  Transactions on Data Science}, vol.~2, no.~3, pp. 1--26, 2021.

\bibitem{bousmalis2018using}
K.~Bousmalis, A.~Irpan, P.~Wohlhart, Y.~Bai, M.~Kelcey, M.~Kalakrishnan,
  L.~Downs, J.~Ibarz, P.~Pastor, K.~Konolige \emph{et~al.}, ``Using simulation
  and domain adaptation to improve efficiency of deep robotic grasping,'' in
  \emph{2018 IEEE international conference on robotics and automation
  (ICRA)}.\hskip 1em plus 0.5em minus 0.4em\relax IEEE, 2018, pp. 4243--4250.

\bibitem{shah2018airsim}
S.~Shah, D.~Dey, C.~Lovett, and A.~Kapoor, ``Airsim: High-fidelity visual and
  physical simulation for autonomous vehicles,'' in \emph{Field and service
  robotics}.\hskip 1em plus 0.5em minus 0.4em\relax Springer, 2018, pp.
  621--635.

\bibitem{coumans2016pybullet}
E.~Coumans and Y.~Bai, ``Pybullet, a python module for physics simulation for
  games, robotics and machine learning,'' 2016.

\bibitem{ehsani2020use}
K.~Ehsani, S.~Tulsiani, S.~Gupta, A.~Farhadi, and A.~Gupta, ``Use the force,
  luke! learning to predict physical forces by simulating effects,'' in
  \emph{Proceedings of the IEEE/CVF Conference on Computer Vision and Pattern
  Recognition}, 2020, pp. 224--233.

\bibitem{mottaghi2016newtonian}
R.~Mottaghi, H.~Bagherinezhad, M.~Rastegari, and A.~Farhadi, ``Newtonian scene
  understanding: Unfolding the dynamics of objects in static images,'' in
  \emph{Proceedings of the IEEE Conference on Computer Vision and Pattern
  Recognition}, 2016, pp. 3521--3529.

\bibitem{tobin2017domain}
J.~Tobin, R.~Fong, A.~Ray, J.~Schneider, W.~Zaremba, and P.~Abbeel, ``Domain
  randomization for transferring deep neural networks from simulation to the
  real world,'' in \emph{2017 IEEE/RSJ international conference on intelligent
  robots and systems (IROS)}.\hskip 1em plus 0.5em minus 0.4em\relax IEEE,
  2017, pp. 23--30.

\bibitem{li2013data}
Y.~Li, J.~Chen, Y.~Zhao, and Q.~Ji, ``Data-free prior model for facial action
  unit recognition,'' \emph{IEEE Transactions on affective computing}, vol.~4,
  no.~2, pp. 127--141, 2013.

\bibitem{teshima2021incorporating}
T.~Teshima and M.~Sugiyama, ``Incorporating causal graphical prior knowledge
  into predictive modeling via simple data augmentation,'' in \emph{Uncertainty
  in Artificial Intelligence}.\hskip 1em plus 0.5em minus 0.4em\relax PMLR,
  2021, pp. 86--96.

\bibitem{Corneanu2019}
C.~Corneanu, M.~Madadi, and S.~Escalera, ``Deep structure inference network for
  facial action unit recognition,'' in \emph{Proceedings of European Conference
  on Computer Vision}, 2019.

\bibitem{chen2019deep}
L.~Chen, H.~Su, and Q.~Ji, ``Deep structured prediction for facial landmark
  detection,'' \emph{Advances in neural information processing systems},
  vol.~32, 2019.

\bibitem{suhail2021energy}
M.~Suhail, A.~Mittal, B.~Siddiquie, C.~Broaddus, J.~Eledath, G.~Medioni, and
  L.~Sigal, ``Energy-based learning for scene graph generation,'' in
  \emph{Proceedings of the IEEE/CVF Conference on Computer Vision and Pattern
  Recognition}, 2021, pp. 13\,936--13\,945.

\bibitem{song2021hybrid}
T.~Song, Z.~Cui, W.~Zheng, and Q.~Ji, ``Hybrid message passing with
  performance-driven structures for facial action unit detection,'' in
  \emph{Proceedings of the IEEE/CVF Conference on Computer Vision and Pattern
  Recognition}, 2021, pp. 6267--6276.

\bibitem{srinivas2021matching}
S.~Srinivas~Kancheti, A.~Gowtham~Reddy, V.~N. Balasubramanian, and A.~Sharma,
  ``Matching learned causal effects of neural networks with domain priors,''
  \emph{arXiv e-prints}, pp. arXiv--2111, 2021.

\bibitem{rieger2020interpretations}
L.~Rieger, C.~Singh, W.~Murdoch, and B.~Yu, ``Interpretations are useful:
  penalizing explanations to align neural networks with prior knowledge,'' in
  \emph{International conference on machine learning}.\hskip 1em plus 0.5em
  minus 0.4em\relax PMLR, 2020, pp. 8116--8126.

\bibitem{zhang2018classifier}
Y.~Zhang, W.~Dong, B.-G. Hu, and Q.~Ji, ``Classifier learning with prior
  probabilities for facial action unit recognition,'' in \emph{Proceedings of
  the IEEE Conference on Computer Vision and Pattern Recognition}, 2018, pp.
  5108--5116.

\bibitem{cui2020knowledge}
Z.~Cui, T.~Song, Y.~Wang, and Q.~Ji, ``Knowledge augmented deep neural networks
  for joint facial expression and action unit recognition,'' \emph{Advances in
  Neural Information Processing Systems}, vol.~33, 2020.

\bibitem{yang2021causalvae}
M.~Yang, F.~Liu, Z.~Chen, X.~Shen, J.~Hao, and J.~Wang, ``Causalvae:
  Disentangled representation learning via neural structural causal models,''
  in \emph{Proceedings of the IEEE/CVF Conference on Computer Vision and
  Pattern Recognition}, 2021, pp. 9593--9602.

\bibitem{qu2019probabilistic}
M.~Qu and J.~Tang, ``Probabilistic logic neural networks for reasoning,''
  \emph{arXiv preprint arXiv:1906.08495}, 2019.

\bibitem{xu2018semantic}
J.~Xu, Z.~Zhang, T.~Friedman, Y.~Liang, and G.~Broeck, ``A semantic loss
  function for deep learning with symbolic knowledge,'' in \emph{International
  conference on machine learning}.\hskip 1em plus 0.5em minus 0.4em\relax PMLR,
  2018, pp. 5502--5511.

\bibitem{xie2019embedding}
Y.~Xie, Z.~Xu, M.~S. Kankanhalli, K.~S. Meel, and H.~Soh, ``Embedding symbolic
  knowledge into deep networks,'' \emph{arXiv preprint arXiv:1909.01161}, 2019.

\bibitem{liang2018symbolic}
X.~Liang, Z.~Hu, H.~Zhang, L.~Lin, and E.~P. Xing, ``Symbolic graph reasoning
  meets convolutions,'' \emph{Advances in Neural Information Processing
  Systems}, vol.~31, pp. 1853--1863, 2018.

\bibitem{liu2021auto}
F.~Liu, C.~You, X.~Wu, S.~Ge, X.~Sun \emph{et~al.}, ``Auto-encoding knowledge
  graph for unsupervised medical report generation,'' \emph{Advances in Neural
  Information Processing Systems}, vol.~34, 2021.

\bibitem{luo2020context}
R.~Luo, N.~Zhang, B.~Han, and L.~Yang, ``Context-aware zero-shot recognition,''
  in \emph{Proceedings of the AAAI Conference on Artificial Intelligence},
  vol.~34, no.~07, 2020, pp. 11\,709--11\,716.

\bibitem{fang2017object}
Y.~Fang, K.~Kuan, J.~Lin, C.~Tan, and V.~Chandrasekhar, ``Object detection
  meets knowledge graphs.''\hskip 1em plus 0.5em minus 0.4em\relax
  International Joint Conferences on Artificial Intelligence, 2017.

\bibitem{gu2019scene}
J.~Gu, H.~Zhao, Z.~Lin, S.~Li, J.~Cai, and M.~Ling, ``Scene graph generation
  with external knowledge and image reconstruction,'' in \emph{Proceedings of
  the IEEE Conference on Computer Vision and Pattern Recognition}, 2019, pp.
  1969--1978.

\bibitem{marino2021krisp}
K.~Marino, X.~Chen, D.~Parikh, A.~Gupta, and M.~Rohrbach, ``Krisp: Integrating
  implicit and symbolic knowledge for open-domain knowledge-based vqa,'' in
  \emph{Proceedings of the IEEE/CVF Conference on Computer Vision and Pattern
  Recognition}, 2021, pp. 14\,111--14\,121.

\bibitem{cui2021type}
Z.~Cui, P.~Kapanipathi, K.~Talamadupula, T.~Gao, and Q.~Ji, ``Type-augmented
  relation prediction in knowledge graphs,'' in \emph{Proceedings of the AAAI
  Conference on Artificial Intelligence}, vol.~35, no.~8, 2021, pp. 7151--7159.

\bibitem{kaur2022trustworthy}
D.~Kaur, S.~Uslu, K.~J. Rittichier, and A.~Durresi, ``Trustworthy artificial
  intelligence: a review,'' \emph{ACM Computing Surveys (CSUR)}, vol.~55,
  no.~2, pp. 1--38, 2022.

\bibitem{wang2020physics}
Z.~Wang, W.~Xing, R.~Kirby, and S.~Zhe, ``Physics informed deep kernel
  learning,'' 2020.

\bibitem{sultan2018transferable}
M.~M. Sultan, H.~K. Wayment-Steele, and V.~S. Pande, ``Transferable neural
  networks for enhanced sampling of protein dynamics,'' \emph{Journal of
  chemical theory and computation}, vol.~14, no.~4, pp. 1887--1894, 2018.

\bibitem{unke2021se}
O.~T. Unke, M.~Bogojeski, M.~Gastegger, M.~Geiger, T.~Smidt, and K.-R.
  M{\"u}ller, ``Se (3)-equivariant prediction of molecular wavefunctions and
  electronic densities,'' \emph{arXiv preprint arXiv:2106.02347}, 2021.

\bibitem{yuan2021simpoe}
Y.~Yuan, S.-E. Wei, T.~Simon, K.~Kitani, and J.~Saragih, ``Simpoe: Simulated
  character control for 3d human pose estimation,'' in \emph{Proceedings of the
  IEEE/CVF Conference on Computer Vision and Pattern Recognition}, 2021, pp.
  7159--7169.

\bibitem{shimada2021neural}
S.~Shimada, V.~Golyanik, W.~Xu, P.~P{\'e}rez, and C.~Theobalt, ``Neural
  monocular 3d human motion capture with physical awareness,'' \emph{ACM
  Transactions on Graphics (TOG)}, vol.~40, no.~4, pp. 1--15, 2021.

\bibitem{anderson1972more}
P.~W. Anderson, ``More is different,'' \emph{Science}, vol. 177, no. 4047, pp.
  393--396, 1972.

\bibitem{born2013principles}
M.~Born and E.~Wolf, \emph{Principles of optics: electromagnetic theory of
  propagation, interference and diffraction of light}.\hskip 1em plus 0.5em
  minus 0.4em\relax Elsevier, 2013.

\bibitem{oren1995generalization}
M.~Oren and S.~K. Nayar, ``Generalization of the lambertian model and
  implications for machine vision,'' \emph{International Journal of Computer
  Vision}, vol.~14, no.~3, pp. 227--251, 1995.

\bibitem{kajiya1986rendering}
J.~T. Kajiya, ``The rendering equation,'' in \emph{Proceedings of the 13th
  annual conference on Computer graphics and interactive techniques}, 1986, pp.
  143--150.

\bibitem{cao2018sparse}
X.~Cao, Z.~Chen, A.~Chen, X.~Chen, S.~Li, and J.~Yu, ``Sparse photometric 3d
  face reconstruction guided by morphable models,'' in \emph{Proceedings of the
  IEEE Conference on Computer Vision and Pattern Recognition}, 2018, pp.
  4635--4644.

\bibitem{wang2003illumination}
H.~Wang, S.~Z. Li, Y.~Wang, and W.~Zhang, ``Illumination modeling and
  normalization for face recognition,'' in \emph{2003 IEEE International SOI
  Conference. Proceedings (Cat. No. 03CH37443)}.\hskip 1em plus 0.5em minus
  0.4em\relax IEEE, 2003, pp. 104--111.

\bibitem{xin2019theory}
S.~Xin, S.~Nousias, K.~N. Kutulakos, A.~C. Sankaranarayanan, S.~G. Narasimhan,
  and I.~Gkioulekas, ``A theory of fermat paths for non-line-of-sight shape
  reconstruction,'' in \emph{Proceedings of the IEEE Conference on Computer
  Vision and Pattern Recognition}, 2019, pp. 6800--6809.

\bibitem{kim2020single}
S.~Kim, Y.~Huo, and S.-E. Yoon, ``Single image reflection removal with
  physically-based training images,'' in \emph{Proceedings of the IEEE/CVF
  Conference on Computer Vision and Pattern Recognition}, 2020, pp. 5164--5173.

\bibitem{zhou2021learning}
C.~Zhou, M.~Teng, Y.~Han, C.~Xu, and B.~Shi, ``Learning to dehaze with
  polarization,'' in \emph{Thirty-Fifth Conference on Neural Information
  Processing Systems}, 2021.

\bibitem{petersen2021learning}
F.~Petersen, C.~Borgelt, H.~Kuehne, and O.~Deussen, ``Learning with algorithmic
  supervision via continuous relaxations,'' in \emph{Thirty-Fifth Conference on
  Neural Information Processing Systems}, 2021.

\bibitem{hartley2003multiple}
R.~Hartley and A.~Zisserman, \emph{Multiple view geometry in computer
  vision}.\hskip 1em plus 0.5em minus 0.4em\relax Cambridge university press,
  2003.

\bibitem{lutter2019deep}
M.~Lutter, C.~Ritter, and J.~Peters, ``Deep lagrangian networks: Using physics
  as model prior for deep learning,'' \emph{arXiv preprint arXiv:1907.04490},
  2019.

\bibitem{satorras2019combining}
V.~G. Satorras, Z.~Akata, and M.~Welling, ``Combining generative and
  discriminative models for hybrid inference,'' 2019.

\bibitem{Zhang_2022_CVPR}
Y.~Zhang, J.~Sun, X.~He, H.~Fu, R.~Jia, and X.~Zhou, ``Modeling indirect
  illumination for inverse rendering,'' in \emph{Proceedings of the IEEE/CVF
  Conference on Computer Vision and Pattern Recognition (CVPR)}, June 2022, pp.
  18\,643--18\,652.

\bibitem{zeng2020tossingbot}
A.~Zeng, S.~Song, J.~Lee, A.~Rodriguez, and T.~Funkhouser, ``Tossingbot:
  Learning to throw arbitrary objects with residual physics,'' \emph{IEEE
  Transactions on Robotics}, 2020.

\bibitem{mottaghi2016happens}
R.~Mottaghi, M.~Rastegari, A.~Gupta, and A.~Farhadi, ``“what happens if...”
  learning to predict the effect of forces in images,'' in \emph{European
  conference on computer vision}.\hskip 1em plus 0.5em minus 0.4em\relax
  Springer, 2016, pp. 269--285.

\bibitem{wu2015galileo}
J.~Wu, I.~Yildirim, J.~J. Lim, B.~Freeman, and J.~Tenenbaum, ``Galileo:
  Perceiving physical object properties by integrating a physics engine with
  deep learning,'' \emph{Advances in neural information processing systems},
  vol.~28, pp. 127--135, 2015.

\bibitem{todorov2012mujoco}
E.~Todorov, T.~Erez, and Y.~Tassa, ``Mujoco: A physics engine for model-based
  control,'' in \emph{2012 IEEE/RSJ International Conference on Intelligent
  Robots and Systems}.\hskip 1em plus 0.5em minus 0.4em\relax IEEE, 2012, pp.
  5026--5033.

\bibitem{kubelka1931article}
P.~Kubelka and F.~Munk, ``An article on optics of paint layers,'' \emph{Z.
  Tech. Phys}, vol.~12, no. 593-601, pp. 259--274, 1931.

\bibitem{haker2009self}
M.~Haker, M.~B{\"o}hme, T.~Martinetz, and E.~Barth, ``Self-organizing maps for
  pose estimation with a time-of-flight camera,'' in \emph{Workshop on Dynamic
  3D Imaging}.\hskip 1em plus 0.5em minus 0.4em\relax Springer, 2009, pp.
  142--153.

\bibitem{chen2021anatomy}
T.~Chen, C.~Fang, X.~Shen, Y.~Zhu, Z.~Chen, and J.~Luo, ``Anatomy-aware 3d
  human pose estimation with bone-based pose decomposition,'' \emph{IEEE
  Transactions on Circuits and Systems for Video Technology}, 2021.

\bibitem{li2019large}
A.~Li, T.~Luo, Z.~Lu, T.~Xiang, and L.~Wang, ``Large-scale few-shot learning:
  Knowledge transfer with class hierarchy,'' in \emph{Proceedings of the
  IEEE/CVF Conference on Computer Vision and Pattern Recognition}, 2019, pp.
  7212--7220.

\bibitem{yu2017visual}
R.~Yu, A.~Li, V.~I. Morariu, and L.~S. Davis, ``Visual relationship detection
  with internal and external linguistic knowledge distillation,'' in
  \emph{Proceedings of the IEEE international conference on computer vision},
  2017, pp. 1974--1982.

\bibitem{akula2021robust}
A.~Akula, V.~Jampani, S.~Changpinyo, and S.-C. Zhu, ``Robust visual reasoning
  via language guided neural module networks,'' \emph{Advances in Neural
  Information Processing Systems}, vol.~34, 2021.

\bibitem{rytting2021leveraging}
C.~Rytting and D.~Wingate, ``Leveraging the inductive bias of large language
  models for abstract textual reasoning,'' \emph{Advances in Neural Information
  Processing Systems}, vol.~34, 2021.

\bibitem{yao2021refining}
H.~Yao, Y.~Chen, Q.~Ye, X.~Jin, and X.~Ren, ``Refining language models with
  compositional explanations,'' \emph{Advances in Neural Information Processing
  Systems}, vol.~34, 2021.

\bibitem{ekman1978facial}
P.~Ekman and W.~V. Friesen, ``Facial action coding system,''
  \emph{Environmental Psychology \& Nonverbal Behavior}, 1978.

\bibitem{enderton2001mathematical}
H.~B. Enderton, \emph{A mathematical introduction to logic}.\hskip 1em plus
  0.5em minus 0.4em\relax Elsevier, 2001.

\bibitem{Cui2019Image}
C.~{Menglong}, J.~{Detao}, Z.~{Ting}, Z.~{Dehai}, X.~{Cheng}, C.~{Zhibo}, and
  X.~{Xiaoqiang}, ``Image classification based on image knowledge graph and
  semantics,'' in \emph{2019 IEEE 23rd International Conference on Computer
  Supported Cooperative Work in Design (CSCWD)}, 2019, pp. 81--86.

\bibitem{ji2019microsoft}
L.~Ji, Y.~Wang, B.~Shi, D.~Zhang, Z.~Wang, and J.~Yan, ``Microsoft concept
  graph: Mining semantic concepts for short text understanding,'' \emph{Data
  Intelligence}, vol.~1, no.~3, pp. 238--270, 2019.

\bibitem{towell1994knowledge}
G.~G. Towell and J.~W. Shavlik, ``Knowledge-based artificial neural networks,''
  \emph{Artificial intelligence}, vol.~70, no. 1-2, pp. 119--165, 1994.

\bibitem{garcez1999connectionist}
A.~S.~A. Garcez and G.~Zaverucha, ``The connectionist inductive learning and
  logic programming system,'' \emph{Applied Intelligence}, vol.~11, no.~1, pp.
  59--77, 1999.

\bibitem{riegel2020logical}
R.~Riegel, A.~Gray, F.~Luus, N.~Khan, N.~Makondo, I.~Y. Akhalwaya, H.~Qian,
  R.~Fagin, F.~Barahona, U.~Sharma \emph{et~al.}, ``Logical neural networks,''
  \emph{arXiv preprint arXiv:2006.13155}, 2020.

\bibitem{donadello2017logic}
I.~Donadello, L.~Serafini, and A.~D. Garcez, ``Logic tensor networks for
  semantic image interpretation,'' \emph{arXiv preprint arXiv:1705.08968},
  2017.

\bibitem{diligenti2017semantic}
M.~Diligenti, M.~Gori, and C.~Sacca, ``Semantic-based regularization for
  learning and inference,'' \emph{Artificial Intelligence}, vol. 244, pp.
  143--165, 2017.

\bibitem{lehmann2015dbpedia}
J.~Lehmann, R.~Isele, M.~Jakob, A.~Jentzsch, D.~Kontokostas, P.~N. Mendes,
  S.~Hellmann, M.~Morsey, P.~Van~Kleef, S.~Auer \emph{et~al.}, ``Dbpedia--a
  large-scale, multilingual knowledge base extracted from wikipedia,''
  \emph{Semantic web}, vol.~6, no.~2, pp. 167--195, 2015.

\bibitem{speer2017conceptnet}
R.~Speer, J.~Chin, and C.~Havasi, ``Conceptnet 5.5: An open multilingual graph
  of general knowledge,'' in \emph{Thirty-First AAAI Conference on Artificial
  Intelligence}, 2017.

\bibitem{enright2013bayesian}
C.~G. Enright, M.~G. Madden, and N.~Madden, ``Bayesian networks for
  mathematical models: techniques for automatic construction and efficient
  inference,'' \emph{International Journal of Approximate Reasoning}, vol.~54,
  no.~2, pp. 323--342, 2013.

\bibitem{evers2011constructing}
S.~Evers and P.~J. Lucas, ``Constructing bayesian networks for linear dynamic
  systems,'' \emph{BMAW-11 Preface}, 2011.

\end{thebibliography}

\begin{IEEEbiographynophoto}{Zijun Cui} 
received the B.S. degree from
the Department of Physics, University of Science and Technology of China, Hefei, China, in 2015, and received the M.S. degree from
the School of Engineering, Brown University, Rhode Island, USA, in 2017.
She is currently pursuing the Ph.D.
degree with the Rensselaer Polytechnic Institute, NY, USA.
She has broad experiences with deep  learning and probabilistic graphical models. Her current research interests include knowledge-augmented deep learning, learning and inference on probabilistic graphical models, and their applications to computer vision and natural language processing.
\end{IEEEbiographynophoto}

\begin{IEEEbiographynophoto}{Tian Gao}  received the Ph.D and B.S degrees both from
the Department of Electrical, Computer, and System
Engineering, Rensselaer Polytechnic Institute, Troy,
NY, USA. He is currently a research staff member of IBM Research AI at T. J. Watson Research Center. His research focuses on machine learning and its application in computer vision and natural language processing. He has worked on probabilistic graphical models, knowledge extraction, causal discovery, and other aspects of machine learning. From 2010 to 2012, he was a National Science Foundation Triple Helix Program Fellow. 
\end{IEEEbiographynophoto}

\begin{IEEEbiographynophoto}{Kartik Talamadupula} is a Senior Research Scientist and Research Manager at IBM Research AI. His background is in automated planning and sequential decision making. He has applied decision-making techniques to human-in-the-loop AI, data-driven dialog, and human-agent collaboration.
\end{IEEEbiographynophoto}

\begin{IEEEbiographynophoto}{Qiang Ji} 
received his Ph.D degree in Electrical Engineering from the University of Washington. He is currently a Professor with the Department of Electrical, Computer, and Systems Engineering at Rensselaer Polytechnic Institute (RPI).  From 2009 to 2010, he served as a program director at the National Science Foundation (NSF), Arlington, VA, USA, where he managed NSF’s computer vision and machine learning programs.  He also held teaching and research positions with the Beckman Institute at University of Illinois at Urbana-Champaign, Urbana, IL, USA; the Robotics Institute at Carnegie Mellon University, Pittsburgh, PA, USA; the Dept. of Computer Science at University of Nevada, Reno, Nevada, USA; and the Air Force Research Laboratory, Rome, NY, USA.  Prof. Ji currently serves as the director of the Intelligent Systems Laboratory (ISL) at RPI.
        Prof. Ji's research interests are in human-centered computer vision, probabilistic graphical models, probabilistic deep learning, and their applications in various fields.  He has published over 300 papers in peer-reviewed journals and conferences, and has received multiple awards for his work.   Prof.  Ji is has served as an editor on several related IEEE and international journals and as a general chair, program chair, technical area chair, and program committee member for numerous international conferences/workshops.   Prof. Ji is a fellow of the IEEE and the IAPR.   

\end{IEEEbiographynophoto}
\end{document}